\documentclass{article}

\PassOptionsToPackage{numbers}{natbib}
\usepackage[final]{nips_2016}

\usepackage{tikz}
\usetikzlibrary{shapes, arrows, calc, positioning,matrix}
\tikzset{
data/.style={circle, draw, text centered, minimum height=3em ,minimum width = .5em, inner sep = 2pt},
empty/.style={circle, text centered, minimum height=3em ,minimum width = .5em, inner sep = 2pt},
}

\usepackage[utf8]{inputenc} 
\usepackage[T1]{fontenc}    
\usepackage{hyperref}       
\usepackage{url}            
\usepackage{booktabs}       
\usepackage{amsfonts}       
\usepackage{nicefrac}       
\usepackage{microtype}      
\usepackage{tikz}
\usetikzlibrary{positioning}
\usetikzlibrary{arrows}
\usepackage{dsfont}
\usepackage{comment}

\usepackage{times}
\usepackage{graphicx} 
\usepackage{caption}
\usepackage{wrapfig}

\usepackage{algorithm}
\usepackage{algorithmic}

\usepackage{hyperref}

\usepackage{amsmath}
\usepackage{amssymb}

\usepackage[font=footnotesize]{caption,subcaption}
\usepackage{sidecap}

\usepackage{xspace}

\usepackage{color}

\newcommand{\pushright}[1]{\ifmeasuring@#1\else\omit\hfill$\displaystyle#1$\fi\ignorespaces}

\title{On Machine Learning and Structure \\for Mobile Robots}

\author{
  Markus Wulfmeier \\
  University of Oxford\\
  \texttt{markus@robots.ox.ac.uk} \\
}

\begin{document}

\maketitle


\begin{abstract}
Due to recent advances - compute, data, models - the role of learning in autonomous systems has expanded significantly, rendering new applications possible for the first time. While some of the most significant benefits are obtained in the perception modules of the software stack, other aspects continue to rely on known manual procedures based on prior knowledge on geometry, dynamics, kinematics etc. Nonetheless, learning gains relevance in these modules when data collection and curation become easier than manual rule design. Building on this coarse and broad survey of current research, the final sections aim to provide insights into future potentials and challenges as well as the necessity of structure in current practical applications.
\end{abstract} 


 
 
\tableofcontents

\pagebreak


\section{Introduction}\label{sec:introduction}
This survey\footnote{A more colloquial version including the embedded videos can be found under \href{https://markusrw.github.io/articles/tldr-on-ml-and-structure-for-robotics/}{https://markusrw.github.io/articles/tldr-on-ml-and-structure-for-robotics/} with a summary under \href{http://ori.ox.ac.uk/on-learning-and-prior-structure/}{http://ori.ox.ac.uk/on-learning-and-prior-structure/}} provides an informal overview of current challenges and potentials of learning across various tasks of relevance in robotics and automation. In this context, similar to the long-term discussion on how much innate structure is optimal for artificial intelligence, there is the more short-term question of how to merge traditional programming and learning (e.g. described as \emph{\href{https://twitter.com/ylecun/status/683400055533899776}{differentiable programming}} or \emph{\href{https://medium.com/@karpathy/software-2-0-a64152b37c35}{software 2.0}}) for more narrow applications in efficient, robust and safe automation.
The question about structure as beneficial or limiting aspect becomes arguably easier to answer in the context of robotic near-term applications as we can simply \emph{acknowledge our ignorance} (the missing knowledge about what will work best in the future) and focus on the present to benchmark and combine the most efficient and effective directions.

Existing solutions to many tasks in mobile robotics, such as localisation, mapping, or planning, focus on prior knowledge about the structure of our tasks and environments. This may include geometry or kinematic and dynamic models, which therefore have been built into traditional programs. However, recent successes and the flexibility of fairly unconstrained, learned models shift the focus of new academic and industrial projects. Successes in image recognition (ImageNet) as well as triumphs in reinforcement learning (Atari, Go, Chess) inspire like-minded research.

This survey is by no means complete, its purpose is to provide a high-level review with more details to be found in the respective references. While the references only represent a small subset of available work in each field, the overall review demonstrates the  regularities, connections and principles underlying the tasks and common software solutions.

\subsection{The Broad Question}\label{sec:broadquestion}
Recently, discussions have come up about the potential relevance of reinforcement learning for deployable mobile robots. When hearing these questions, it seems easy to reject them as an executive's fear of missing out on another hyped technology. However, it is worth taking a moment to investigate these questions. 

Deep learning predominantly has made its mark regarding applications in the perception pipeline of autonomous systems including pedestrian / car / cyclist / traffic sign detection, semantic segmentation, and other related tasks. While these perception systems heavily rely on learning; localisation, reasoning, and planning modules often continue to be the domain of carefully crafted rules and programs, exploiting geometric priors and intuitions. The design of these systems requires expert knowledge and repeated iteration between testing - in simulation as well as on the real platform - and refinement of hundreds if not thousands of heuristics. 

While for example in the early DARPA challenges, robotic systems nearly completely relied on these structures, the paradigm is starting to shift \citep{decadeafterdarpa}. Given the success in perception tasks, the natural question is: 'what else can we learn from data?'. Discussions about using (reinforcement) learning naturally arise in the context of reducing manual efforts and instead automatically learning decision patterns. The overall question now focusses on the general application of learning in further parts of our pipeline; with RL representing one of the potentially more \emph{high risk, high reward} scenarios.

\begin{figure}
    \centering
    \begin{subfigure}[t]{0.31\textwidth}
        \includegraphics[width=\textwidth,trim={0 2cm 0 2cm},clip]{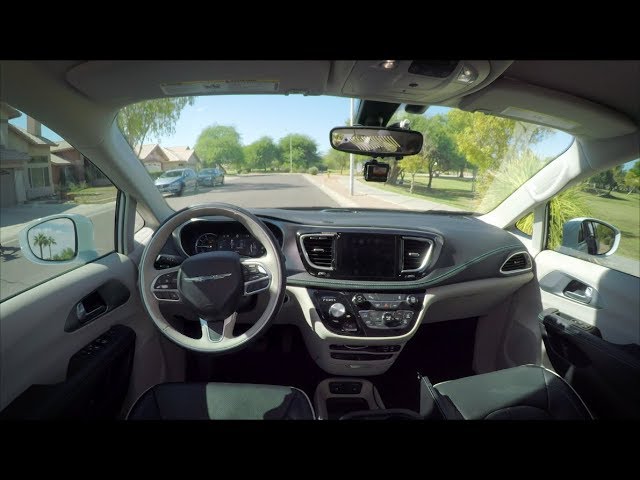}
        \caption{Waymo (\href{https://www.youtube.com/watch?v=tiwVMrTLUWg?rel=0}{link})}
    \end{subfigure} ~
    \begin{subfigure}[t]{0.31\textwidth}
        \includegraphics[width=\textwidth]{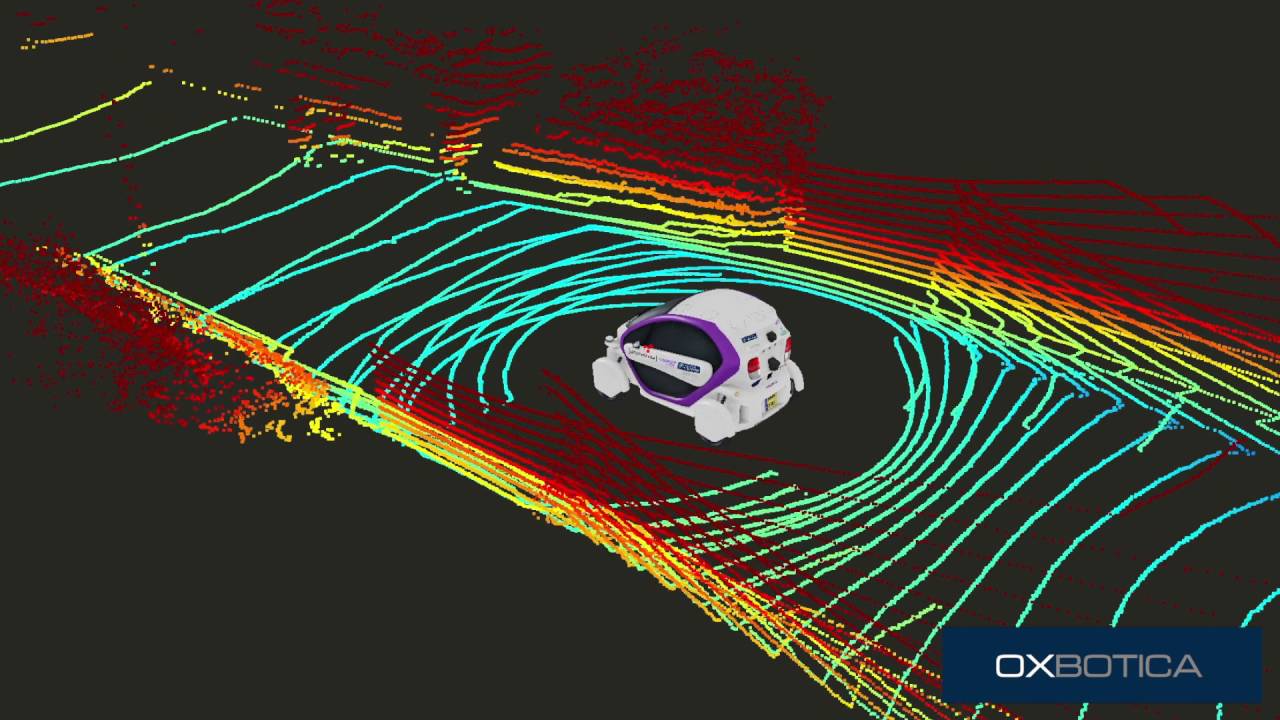}
        \caption{Oxford Robotics Institute / Oxbotica (\href{https://www.youtube.com/watch?v=aaOB-ErYq6Y}{link})}
    \end{subfigure} ~
    \begin{subfigure}[t]{0.31\textwidth}
        \includegraphics[width=\textwidth]{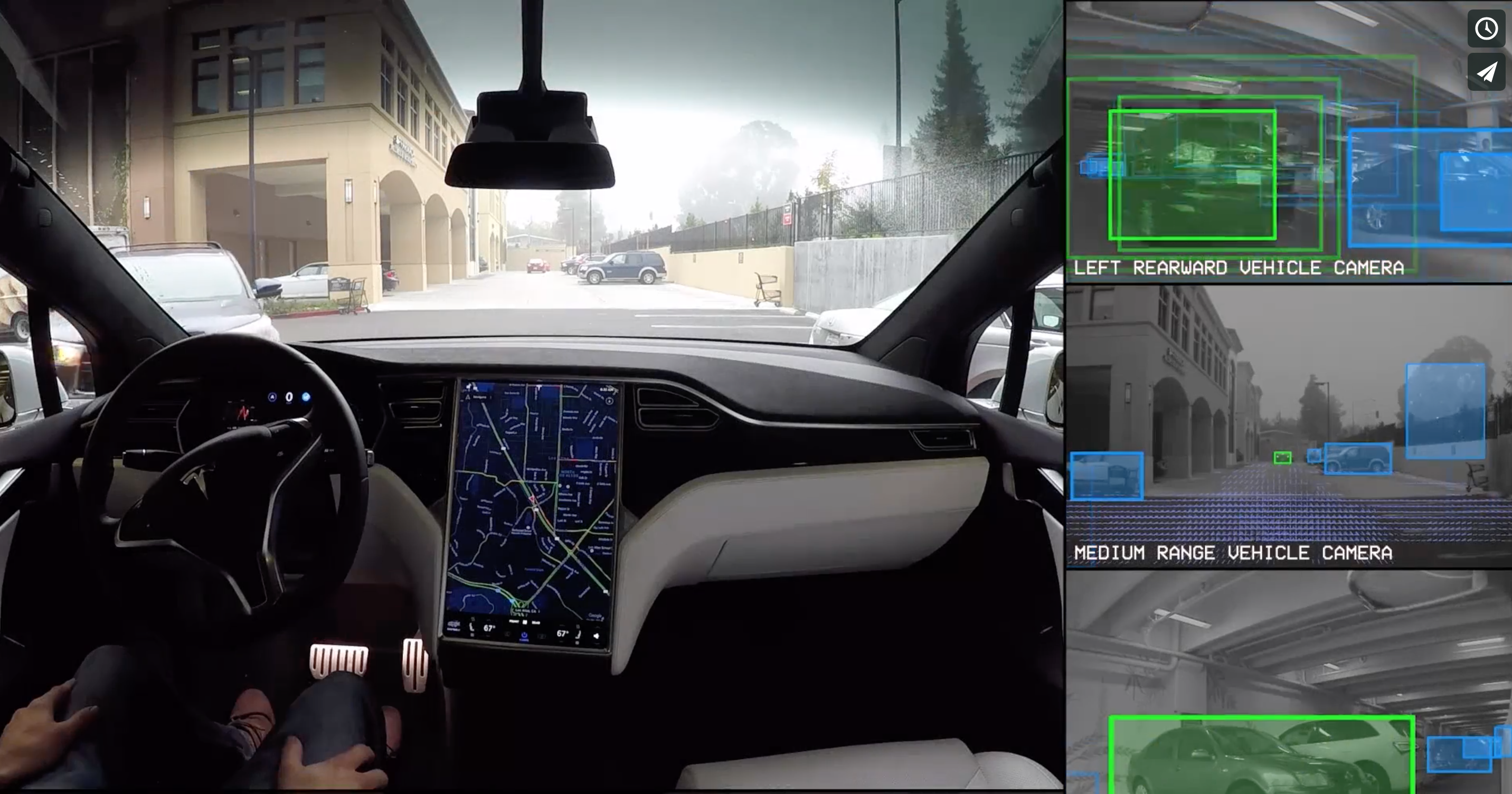}
        \caption{Tesla (\href{https://player.vimeo.com/video/192179727}{link})}
    \end{subfigure}
    
    \vspace{3mm}
    \begin{subfigure}[t]{0.31\textwidth}
        \includegraphics[width=\textwidth,trim={0 2cm 0 2cm},clip]{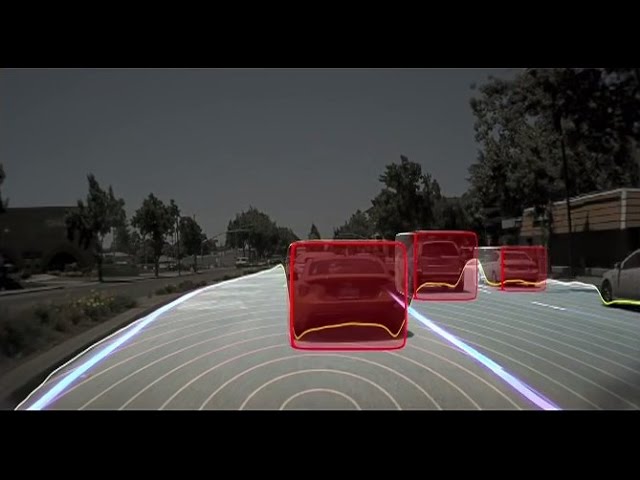}
        \caption{NVIDIA (\href{https://www.youtube.com/watch?v=URmxzxYlmtg?rel=0}{link})}
    \end{subfigure} ~
    \begin{subfigure}[t]{0.31\textwidth}
        \includegraphics[width=\textwidth]{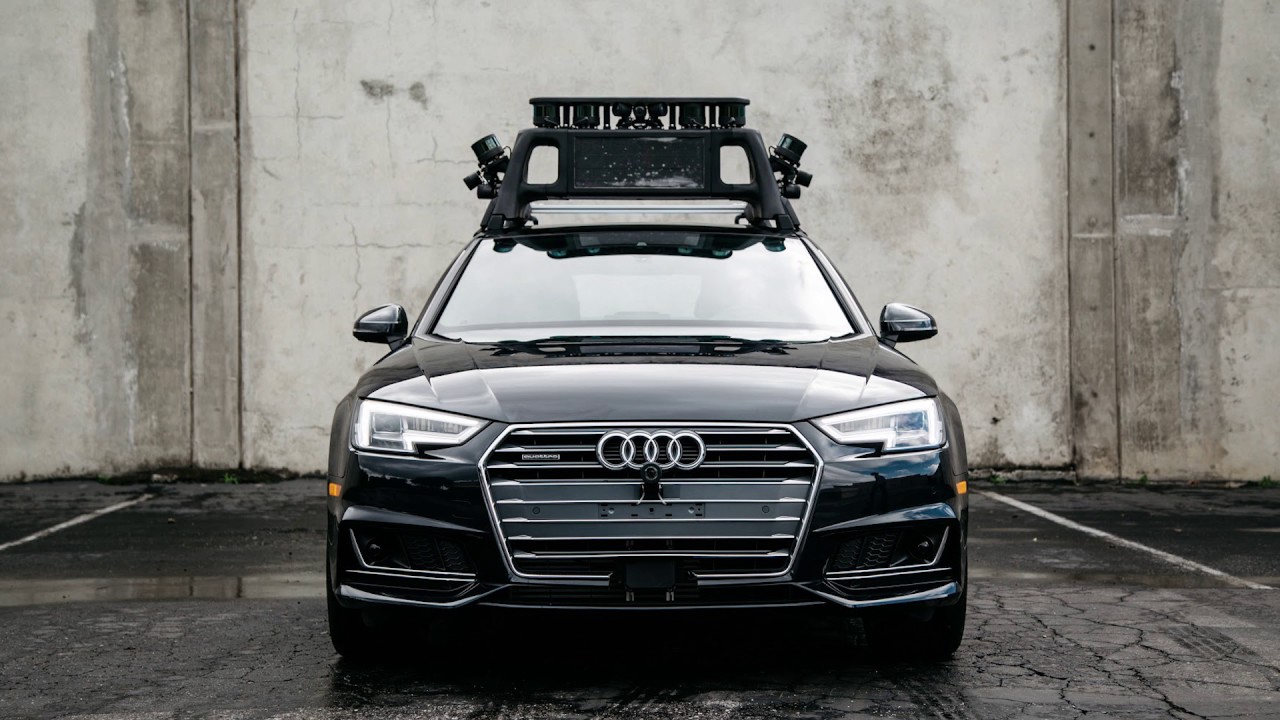}
        \caption{Drive.ai (\href{https://www.youtube.com/watch?v=6Pz5FMmcT-Y?rel=0}{link})}
    \end{subfigure} ~
    \begin{subfigure}[t]{0.31\textwidth}
        \includegraphics[width=\textwidth]{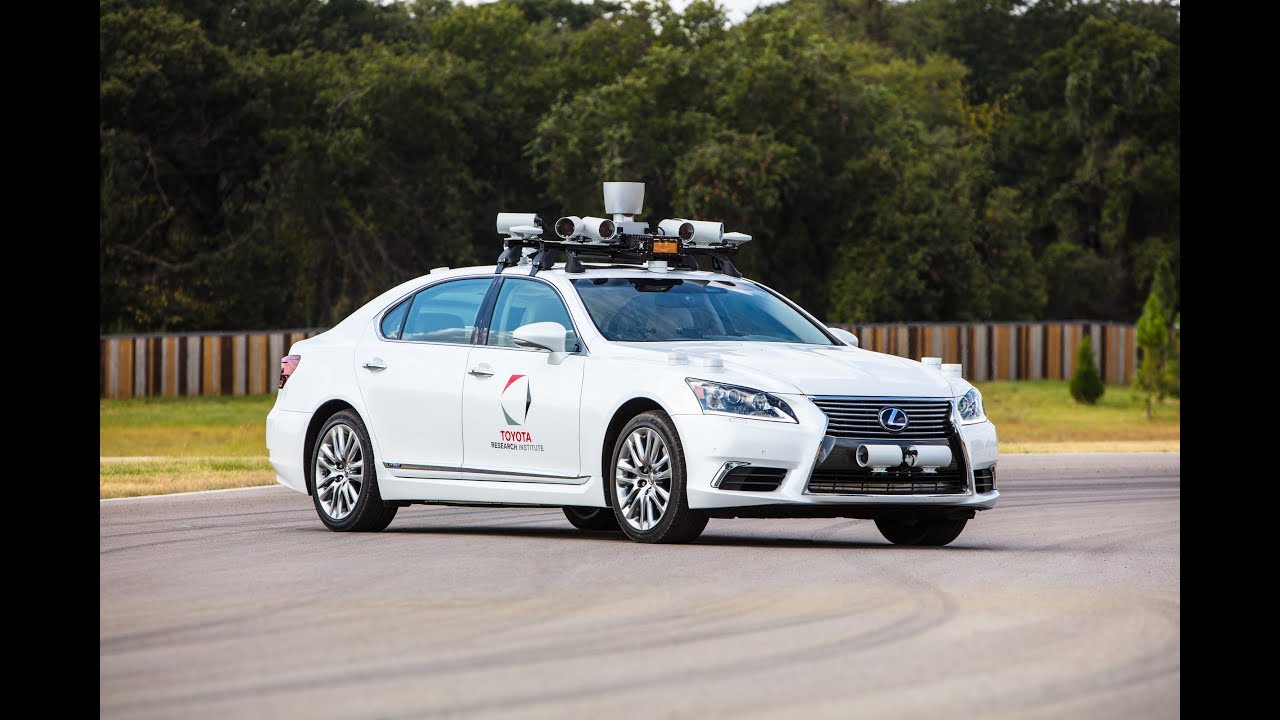}
        \caption{Toyota Research Institute (\href{https://www.youtube.com/watch?v=ajreRfot6co?rel=0}{link})}
    \end{subfigure}
    \label{fig:d}
\end{figure}

While ML has been able to improve our efficiency in addressing various tasks, it does not represent \emph{free lunch}. Independent of all its advantages, machine learning delivers no magic tool. Its successful application commonly requires detailed domain knowledge, systems engineering and demands significant time for data collection and curation, experimental setup and safety arrangements.

\section{Learning for Autonomous Systems}\label{sec:learning}
Autonomous systems are generally modularised for the same reasons as any large software systems: reuseability, ease of testing, separation of responsibilities, interpretability, etc. Robots / autonomous systems are treated in this article as a collection of these modules, including: perception, localisation, mapping, tracking, prediction, planning, and control.

The following paragraphs survey a subsection of work in each field, exemplifying the state of the art of learning based methods for these modules, followed by additional directions of relevance across the whole pipeline on uncertainty and introspection as well as representations.
The following, final sections represent a more personal take on challenges and potentials. More resources on software systems and computer vision for autonomous platforms for interested readers can be found under \citep{apollo, udacity, 2017arXiv170405519J} and generally in the mentioned references.

\subsection{Perception}\label{sec:perception}
Current perception modules represent one of the principal success stories of deep learning in autonomous systems. Image classification, object detection, depth estimation, semantic segmentation, activity recognition are all principally dominated by deep learning \citep{kitti, cityscapes, benchmarks} (a detailed survey of recent work can be found under \citep{themtank}).

While classification benchmarks have been long-standing pillars of computer vision research, the ImageNet benchmark \citep{ILSVRC15} in particular presents a cornerstone for the acceleration of progress in machine learning and in particular deep learning. A good share of models originally was developed specifically for this benchmark. ImageNet dataset as well as benchmark have massive forces for research on deep learning, which triumphed in all recent competitions.

The \textbf{detection} of traffic participants including pedestrians, cyclists and other vehicles \citep{Geiger2012CVPR, cordts2016cityscapes} relies predominantly on deep learning approaches for image \citep{ren2017accurate, 2016arXiv160707155C, 2016arXiv161208242R} as well as LIDAR data \citep{2016arXiv161107759C, 2016arXiv160906666E}. LIDAR can be understood as essentially a light based radar: the sensor's output being a long sequence of distance measurements. Notably, the natural structure of LIDAR significantly differs from image data and is unfit for the application of models designed for images. One essential challenge is that the same pointcloud can be represented by many different sequences and applied models have to be permutation invariant.  
Most early approaches are able to prevail by building on manually designed grids with predefined feature extractors for each cell \citep{2016arXiv160906666E, 2017ISPAn41W191H}. Replacing this kind of manual feature design, a more recent direction is the combination of low-level feature learning based on recurrent modules and high-level grid-based representations via convolutions for end-to-end training \citep{2016arXiv161200593Q}. Further work relies on max-pooling as symmetric function over point-wise descriptors (treating the data as a set rather than as a sequence) and extensions to address local features at varying contextual scales \citep{qi2017pointnet,qi2017pointnetplus}.

\begin{figure}
    \centering
    \begin{subfigure}[t]{0.31\textwidth}
        \includegraphics[width=\textwidth]{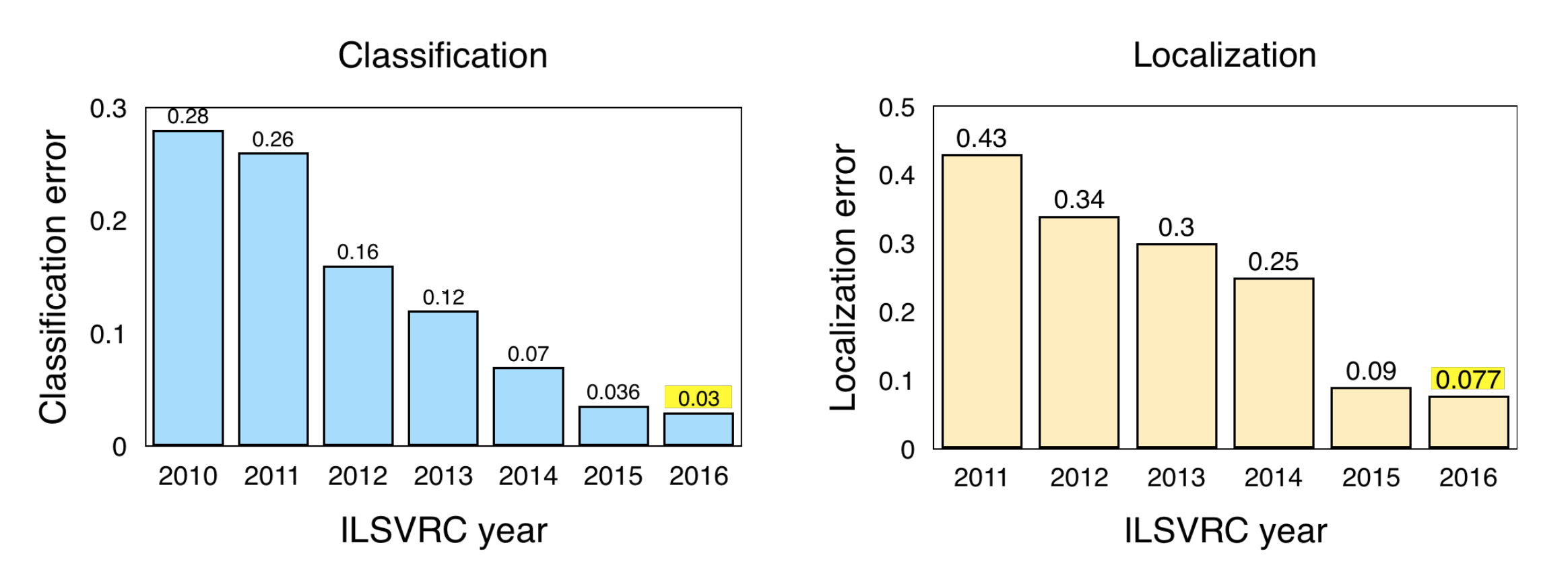}
        \caption{ImageNet (\href{http://image-net.org/challenges/talks/2016/ILSVRC2016_10_09_clsloc.pdf}{link})}
    \end{subfigure} ~
    \begin{subfigure}[t]{0.31\textwidth}
        \includegraphics[width=\textwidth]{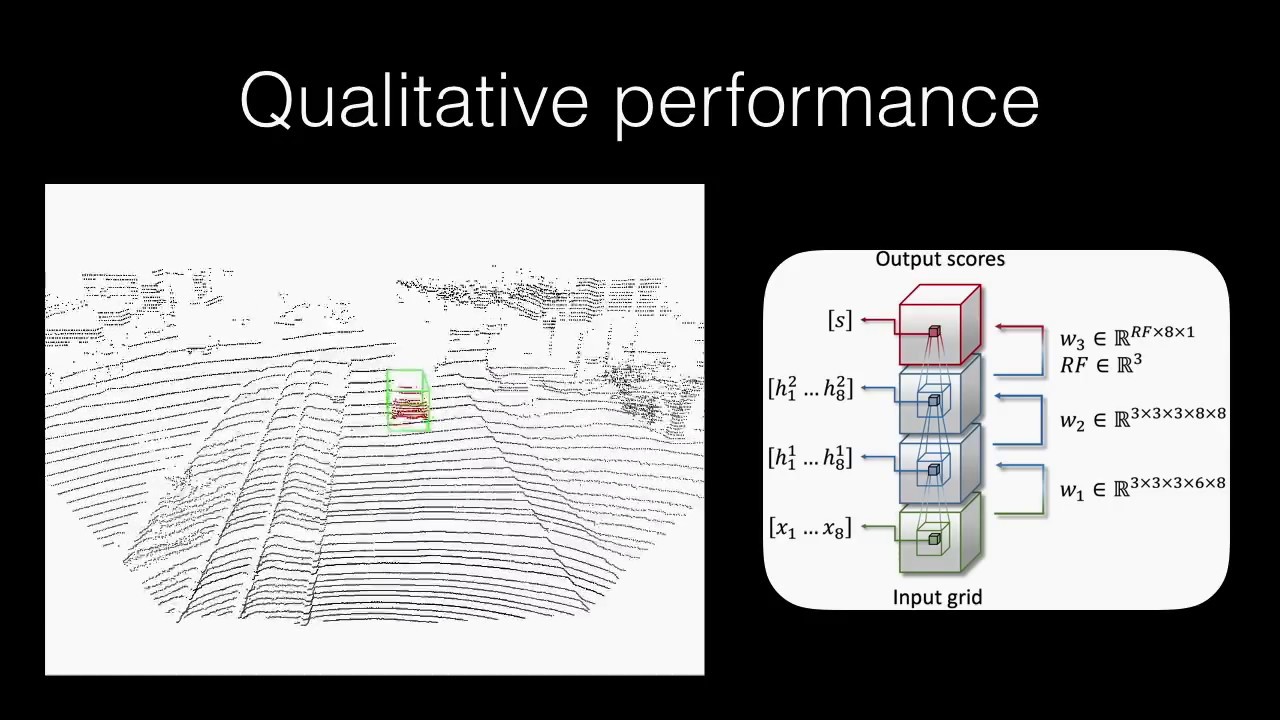}
        \caption{Vote3Deep \citep{2016arXiv160906666E} (\href{https://www.youtube.com/watch?v=WUOSmAfeXIw?rel=0}{link})}
    \end{subfigure} ~
    \begin{subfigure}[t]{0.31\textwidth}
        \includegraphics[width=\textwidth]{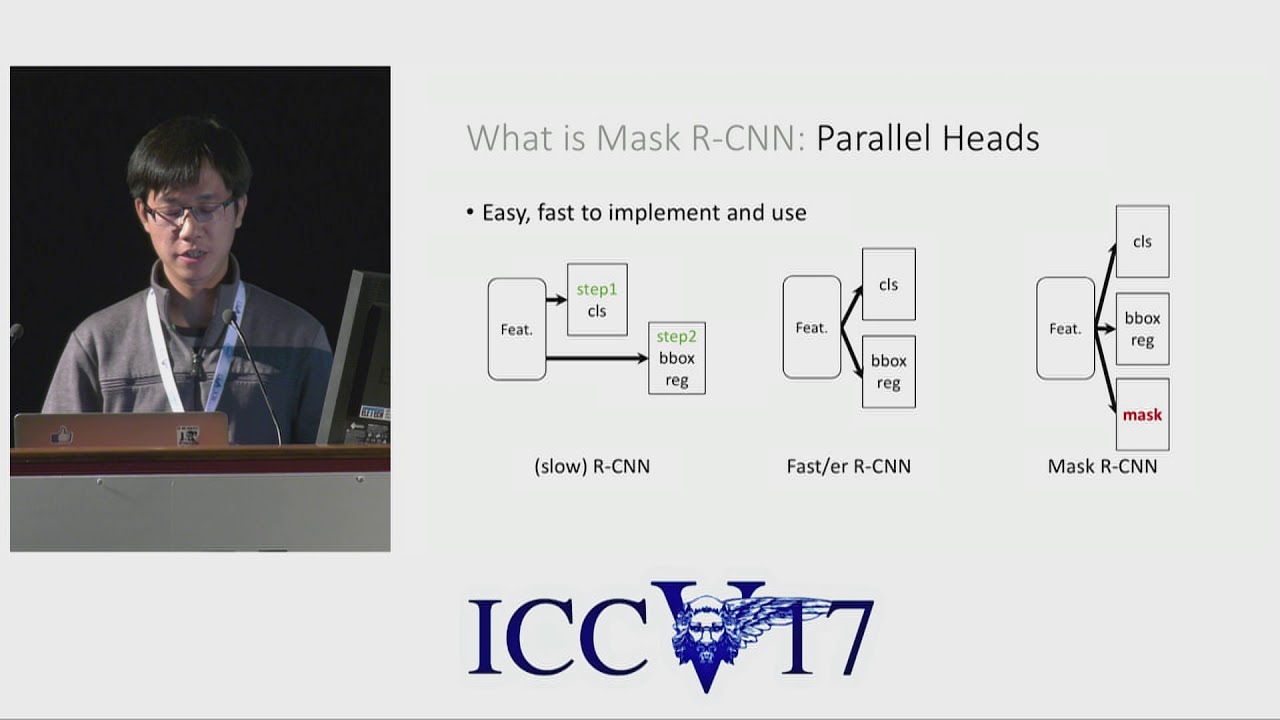}
        \caption{Mask RCNN \citep{2017arXiv170306870H} (\href{https://www.youtube.com/watch?v=g7z4mkfRjI4?rel=0}{link})}
    \end{subfigure}
\end{figure}

Similarly, \textbf{pixel-wise semantic and instance segmentation} \citep{2017arXiv170406857G, 2016arXiv161206370P, cityscapes, kitti} as well as image-based \textbf{depth / disparity estimation} (mono and stereo) \citep{middlebury, kitti, monodepth17, 2016arXiv161202401U, 2017arXiv170202706K, garg2016unsupervised} abide in the domain of deep learning based approaches. Interestingly however, ideas from geometric computer vision are making their way back into current research for the latter direction, e.g. in the form of reprojection losses \citep{monodepth17}. Multi-task training with shared encoder segments has demonstrated additional improvements \citep{2017arXiv170802550N, 2017arXiv170306870H} when parts of the architecture can be shared. Lastly, in general \textbf{scene understanding}, deep learning has been beneficial for tasks such as the prediction of road attributes \citep{2016arXiv161108583S}. 

The applications in this section are predominantly mastered via learning. Innovation often focuses on architecture or loss function design, which exceed the scope of this review. Furthermore, a significant share of intriguing work takes place in product-oriented, more applied research, which is less published.

\subsection{Localisation and Mapping}\label{sec:localisation}

We will start this section by conveying condensed and highly simplified intuitions about the computations underlying current solutions to relative localisation, absolute localisation as well as the full SLAM (simultaneous localisation and mapping) problem.

In relative localisation, we commonly determine matching features in consecutive sensor measurements and, based on the changes in their coordinates, we compute our change in pose; the latter being accurately described via geometric rules (\href{https://en.wikipedia.org/wiki/Camera_matrix}{projection}, \href{https://en.wikipedia.org/wiki/Triangulation_(surveying)}{triangulation}, etc).
Absolute localisation (localisation against a map) additionally involves a feature-matching process of current perception against locations in our map to determine the coarse location and potentially re-localise.
For SLAM, we additionally build a map. While we determine our own location from the estimated positions of features, we estimate the position of new features with respect to the map to update and enhance. Additional refinement of the map for this joint optimisation problem is commonly formulated via \href{https://en.wikipedia.org/wiki/Kalman_filter}{iterative filtering} and \href{https://en.wikipedia.org/wiki/Bundle_adjustment}{bundle adjustment} techniques. The overall problem has many unmentioned challenges based on the efficient, robust realisation of these sub-tasks as well as in the complexity of real-world data including sensor noise, occlusions and dynamic environments.

Applications in localisation and mapping have provided challenging benchmarks for learning-based approaches. Geometric methods e.g. for \textbf{visual odometry}  continue to outperform \emph{end-to-end learning} \citep{kitti} (end-to-end indicating here the learning of the complete odometry image-pose pipeline - in opposition to learning modules such as interest point descriptors \citep{2017arXiv170707410D}). Geometric methods have the benefit of incorporating our prior knowledge about exact geometric rules (e.g. regarding \href{https://en.wikipedia.org/wiki/Homography_(computer_vision)}{homographies} and projections), which learning based methods will at best learn to approximate. 
While we are able to formulate exact equations e.g. for homography estimation, there are various tasks which commonly are solved more heuristically. The sub-optimal compression of available information, as in the context of feature descriptors, provides an opportunity for learning to minimise the loss of relevant information.

However, the actual gap between distinctly geometric or learned approaches for localisation is decreasing in practice due to consolidations of both directions. Recent work combines the flexibility of learned sub-systems and prior knowledge about task-dependent computations incorporating prior intuition from geometric CV \citep{2017arXiv170510279P, 2017arXiv170908429W, 2017arXiv170806822T, 2016arXiv161003344C}. 
One example is given by the integration of auxiliary training losses to address the common drift problem of relative pose estimation \citep{nister2004visual}. 
Additionally to predicting accurate relative transforms, this objective can be applied to the integrated motion over multiple steps \citep{2017arXiv170510279P} to reduce accumulated drift. Furthermore, learning-based approaches provide the benefit of being independent of knowledge about (intrinsic camera) calibration as distorted images can be directly used \citep{kendall2015posenet}.

\textbf{Absolute localisation} - relative to map instead of relative to our last position - commonly relies on such a map populated with features to localise against. Generally, in the context of current deep learning, most approaches utilise no explicit constraint on the type of computation \citep{kendall2015posenet, Brachmann_2016_CVPR}, there are however notable exceptions \citep{2017arXiv170609520Z}). 
Recent work aims at harnessing geometric prior knowledge to obtain more informative training objectives \citep{kendall2017geometric} (with a survey under \citep{geometric_losses}).

\begin{figure}
    \centering
    \begin{subfigure}[t]{0.31\textwidth}
        \includegraphics[width=\textwidth]{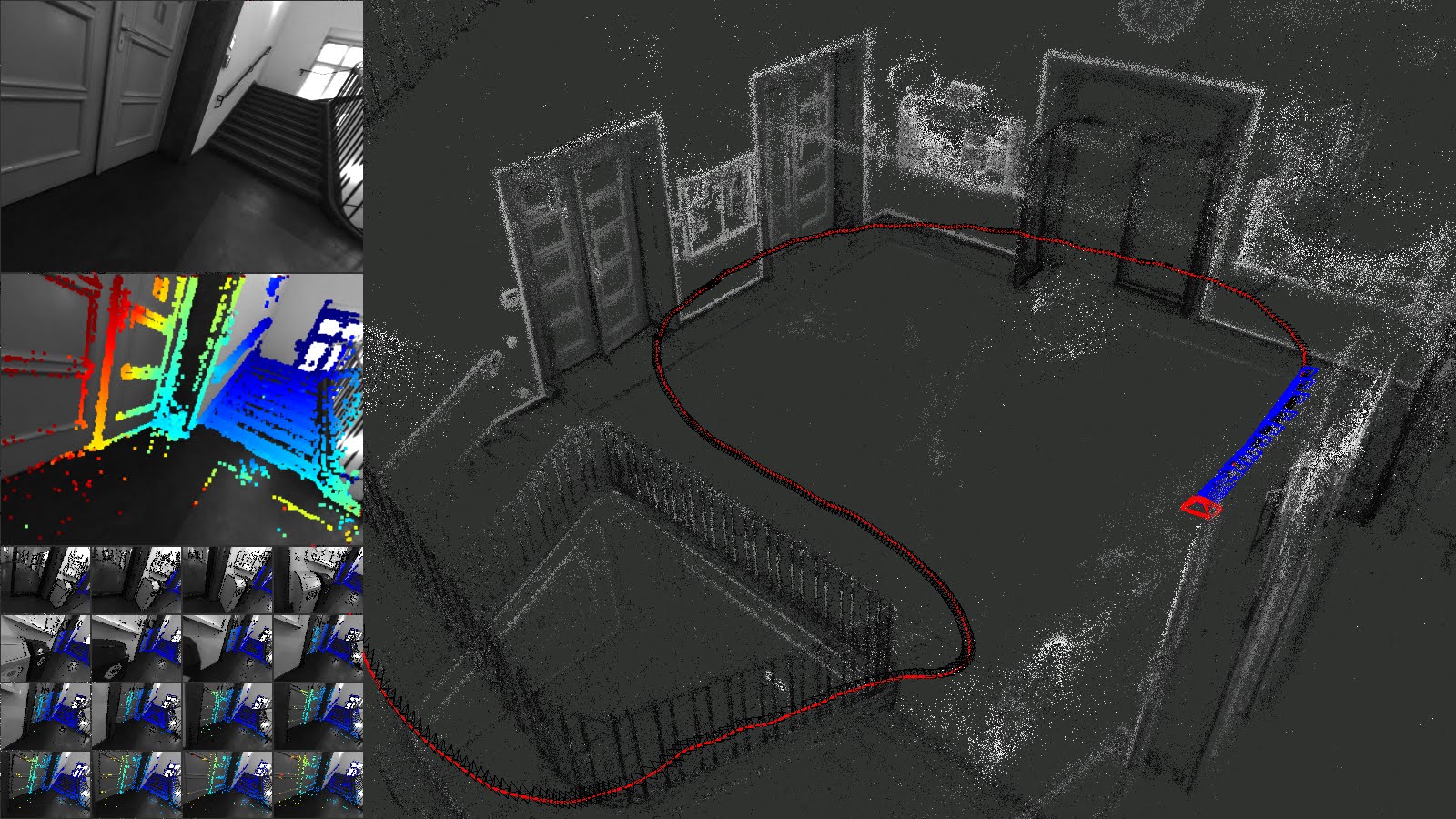}
        \caption{Direct Sparse Odometry \citep{engel2017direct} (\href{https://www.youtube.com/watch?v=C6-xwSOOdqQ?rel=0}{link})}
    \end{subfigure} ~
    \begin{subfigure}[t]{0.31\textwidth}
        \includegraphics[width=\textwidth]{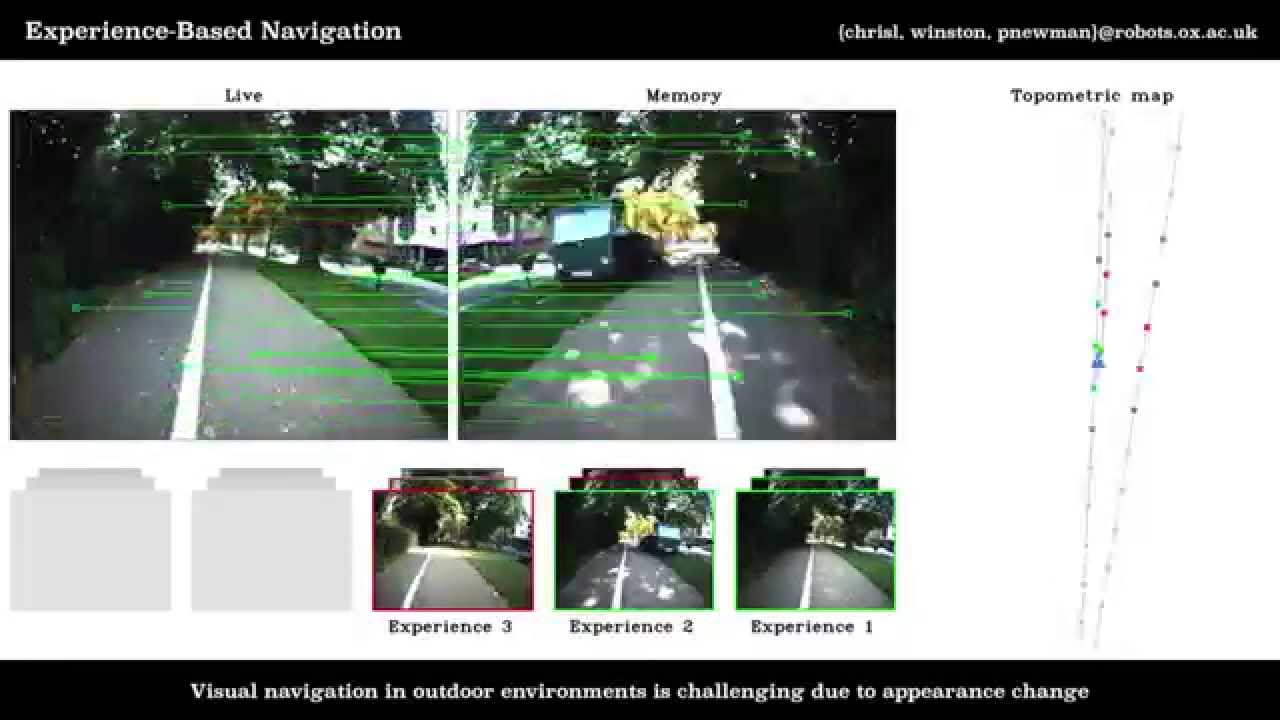}
        \caption{Experience-Based Navigation \citep{churchill2013experience} (\href{https://www.youtube.com/watch?v=4rDy5D1agRw?rel=0}{link})}
    \end{subfigure} ~
    \begin{subfigure}[t]{0.31\textwidth}
        \includegraphics[width=\textwidth]{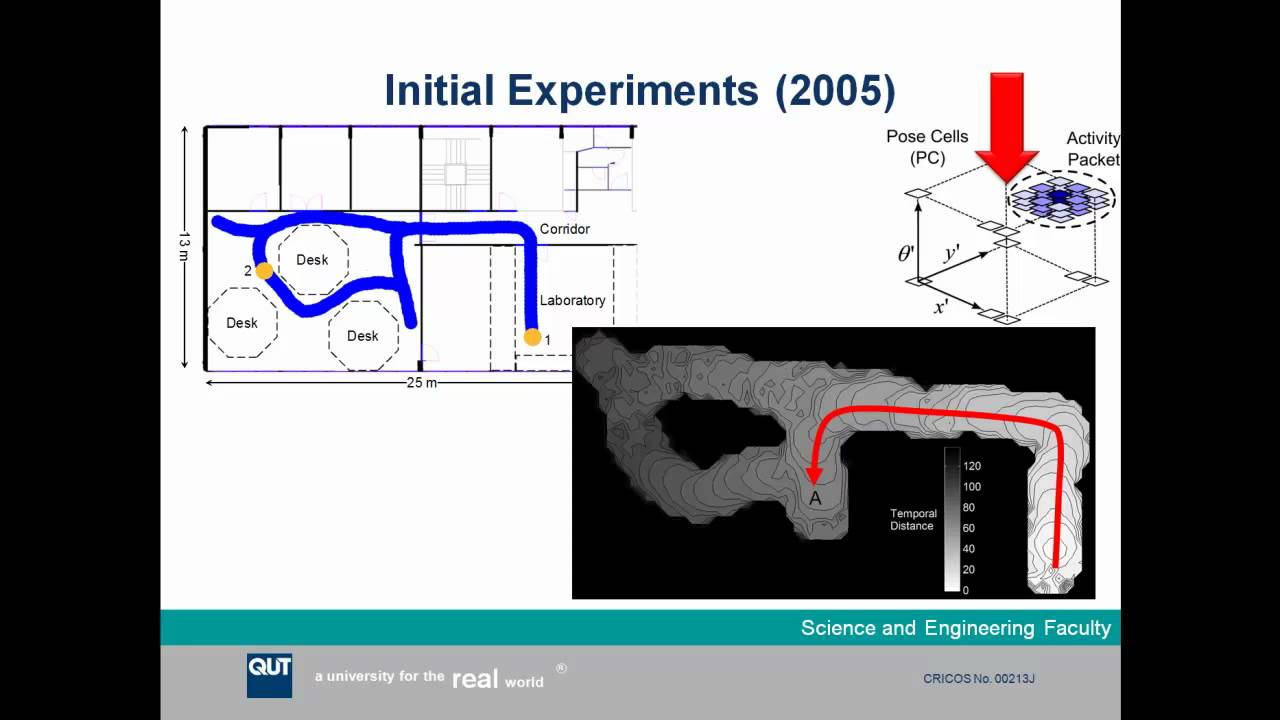}
        \caption{RatSLAM \citep{milford2004ratslam} (\href{https://www.youtube.com/watch?v=t2w6kYzTbr8?rel=0}{link})}
    \end{subfigure}
\end{figure}

Early work on \emph{neural approaches} to the full \textbf{SLAM} problem \citep{slam} is given by Milford and Wyeth \citep{milford2004ratslam} taking inspiration from computational models of the hippocampus of rodents (RatSLAM). More recently, Tateno and colleagues \citep{tateno2017cnn} apply learning to solve a sub-problem of SLAM: using a convolutional neural network as depth estimator (see Section \ref{sec:perception}) to overcome shortcomings of monocular SLAM regarding absolute scale. Another approach to \emph{neural SLAM} is taken in \citep{2017arXiv170707410D} bu addressing another task within the SLAM systems and providing a fast deep learning based point tracking systems. An extension of their work (and combination with homography estimation \citep{2016arXiv160603798D}) extends past synthetic data and outperforms various learned and non-learned point detector/descriptor baselines on a range of tasks \citep{2017arXiv171207629D}.
A final approach to neural SLAM is given by Zhang et al \citep{2017arXiv170609520Z}, who try to \emph{embed procedures mimicking that of traditional Simultaneous Localization and Mapping (SLAM) into the soft attention-based addressing of external memory architectures, in which the external memory acts as an internal representation of the environment}. In essence, the authors aim at providing a model, which inherently encourages learning SLAM-like procedures. However, the evaluation focuses on limited toy examples in simulation.

To enable loop closures, \textbf{place recognition} addresses the recognition of previously visited locations based on their appearance and is a relevant part of the SLAM pipeline. This represents a general localisation sub-problem well-suited for learning-based approaches \citep{ arandjelovic2016netvlad}. Commonly, it is targeted in the context of comparing the feature representations between potential candidates \citep{sunderhauf2015place, arandjelovic2016netvlad} (the latter being a differentiable adaptation of the VLAD image descriptor \citep{jegou2010aggregating}). Finally, appearance change in our environment continues to be one of the most challenging aspects for learning as well as geometric approaches \citep{lowry2016visual}. Work on obtaining weather or lighting invariant representations to address this challenge is summarised in Section \ref{sec:transfer}. 

Predominantly, current combined methods for localisation focus on employing learning for sub-tasks which are only heuristically solved in traditional CV as well as utilising geometrically inspired structure and computations to \emph{learn into}.

\subsection{Tracking and Prediction}\label{sec:tracking}
In essence, most object tracking pipelines can be divided into two steps: prediction of all tracks based on a prediction model and creation as well as update of the tracks based on current measurements; the latter depending on accurate assignments between current measurements and existing tracks. 

Focusing on multi-object scenarios, one of the principal challenges of tracking lies in the \textbf{data association} problem: knowing which of the current detections corresponds to which established track. The most common methods for high-frequency tracking pipelines rely on simple distance-based associations between predicted position and current detections. 
However, in cluttered environments and the context of occlusions, additional information such as appearance is required to enable accurate tracking. Learning-based approaches have been successfully applied here e.g. to metric learning for appearance-based entity re-identification for pedestrians \citep{Liao_2015_CVPR}. By applying objectives such as contrastive \citep{hadsell2006dimensionality}, triplet \citep{hoffer2015deep} and magnet loss \citep{rippel2015metric} these methods learn a metric space where different instances of the same type reside closer together. Further methods train models for direct appearance-based tracking \citep{kristan2016visual, bertinetto2016fully} (more details in Section \ref{sec:bottomup}).

To associate existing tracks with new detections or provide position updates at potentially higher frequency than the received detections, we need to predict future positions. A classic, the (extended/unscented) Kalman filter \citep{julier1997new, wan2000unscented}, is actually sufficient in most common situations. The simple \textbf{motion models} often underlying these methods (e.g. constant velocities) though will turn out unreliable in the context of long-term predictions and cluttered environments. For accurate \textbf{prediction} of future trajectories, more information such as interactions with static scenery as well as other agents (cyclists, cars, pedestrians) needs to be considered. 

Early work to represent these interactions with \emph{social force} models \citep{helbing1995social} applies potential fields for modelling repulsive and attracting forces. Reciprocal Velocity Obstacles (RVO) were introduced as a computationally efficient extension with applications not only directly for prediction but also integrated into motion planning \citep{van2008reciprocal, AlonsoMora2013} though the approach requires additional knowledge about the interacting entities. 
Recent work on learned predictive models employs deep neural networks to integrate information about static environment and dynamic environments \citep{ alahi2016social, 2017arXiv170205552F, 2017arXiv170908528P}. 
In addition to flexibly utilising large quantities of raw sensor measurements, these approaches have been shown to be able to partially address the drift of trajectories by directly predicting multi-step sequences \citep{2017arXiv170908528P}. 
Notably, the KF itself has become a target for learning. BackprobKF \citep{2016arXiv160507148H} provides a fully differentiable architecture for state estimation which is evaluated on the KITTI visual odometry benchmark \citep{kitti}. 

\begin{figure}
    \centering
    \begin{subfigure}[t]{0.31\textwidth}
        \includegraphics[width=\textwidth]{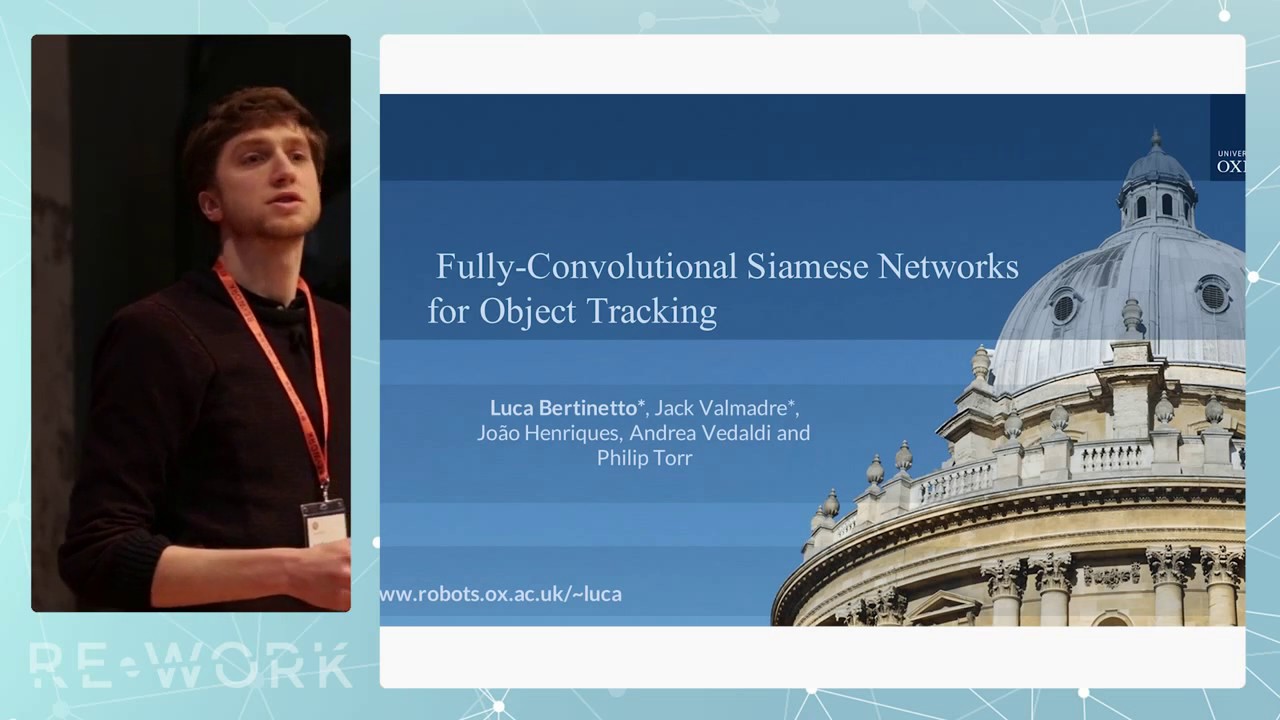}
        \caption{Fully Convolutional Siamese Networks for Object Tracking \citep{bertinetto2016fully} (\href{https://www.youtube.com/watch?v=jZoUalMMZ_0?rel=0}{link})}
    \end{subfigure} ~
    ~ 
    \begin{subfigure}[t]{0.31\textwidth}
        \includegraphics[width=\textwidth, clip, trim = {0 2cm 0 2cm}]{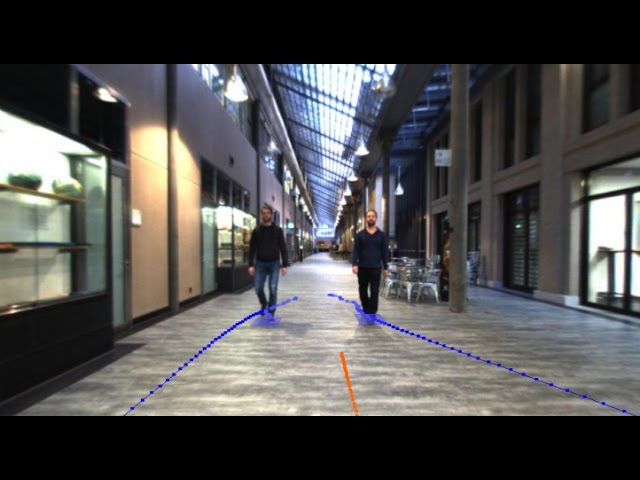}
        \caption{Predict Actions to Act Predictably \citep{PfeifferIROS2016} (\href{https://www.youtube.com/watch?v=h1rm0BW3eVE?rel=0}{link})}
    \end{subfigure} ~
    ~
    \begin{subfigure}[t]{0.31\textwidth}
        \includegraphics[width=\textwidth]{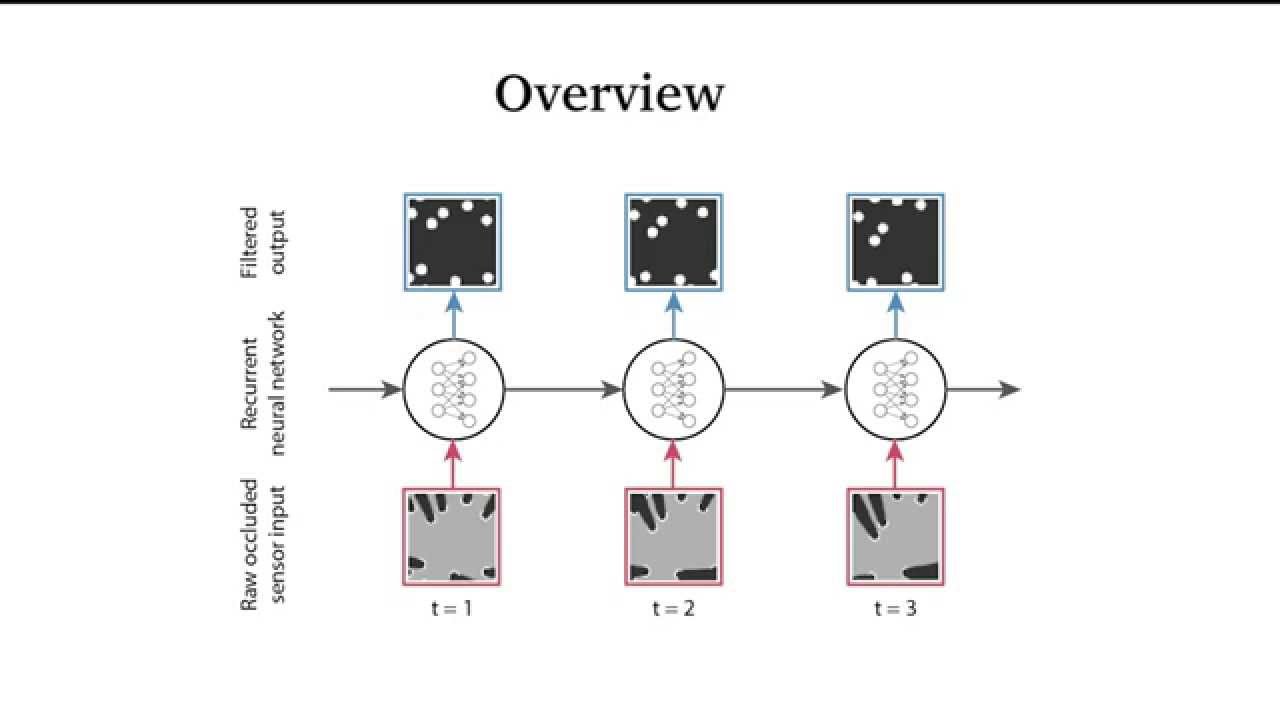}
        \caption{Deep Tracking (precursor to \citep{dequaire2017deep}) (\href{https://www.youtube.com/watch?v=cdeWCpfUGWc?rel=0}{link})}
    \end{subfigure}
\end{figure}

Given the perspective of robotics, we're often interested not just in the prediction of tracks for known and detected objects but the complete prediction of future states ,including aspects we do net explicitly handle in the detection module. 
Addressing this challenge, a different angle to tracking is given by approaches like 'Deep Tracking' \citep{dequaire2017deep} which predict complete future sensor observations (e.g. LIDAR and camera) \citep{finn2016unsupervised, mathieu2015deep, kalchbrenner2016video, vondrick2017generating, 2017arXiv170508781H}. These methods bypass the data association problem as well as the general detection challenge and can provide redundancy for the prediction of future observations, such as occupancy grids. 
However, learning generative models for the prediction of complete sensor measurements has so far proven particularly challenging. 

In general, the prediction of future motion, in particular other agents' reactions, has great benefits for the following modules including motion planning \citep{2017arXiv171009483S, PfeifferIROS2016}.

\subsection{Planning and Control}\label{sec:planning}
Planning and control are the final components of our pipeline and the connecting modules to determine commands for actuation. A principal question for these modules is the type and source of supervision. While a significant share of currently deployed solutions builds on manually hand-crafted rules, learning provides a relevant alternative to prevent repeated hyperparameter and heuristic tuning for different environments and scenarios. Now, one solution for supervision signal can be through reinforcement learning, which - while representing a multi-faceted topic of its own - still needs to overcome many \href{https://www.alexirpan.com/2018/02/14/rl-hard.html}{real-world challenges} and simply is to intricate to cover as just a side aspect of this review. This section mostly focuses on \textbf{Learning from Demonstration} to provide supervision based on demonstrations of a task from human experts and other potential authorities. 

\textbf{Behavioural Cloning} (BC) aims at directly mimicking expert behaviour to solve a task; essentially supervised optimisation of regression or classification models. BC can be integrated into existing pipelines to build on more abstract representations but most commonly has been investigated in the scenario of end-to-end learning based on raw inputs. These methods have been empirically demonstrated in constrained scenarios e.g. for lane keeping \citep{bojarski2016end, pomerleau1990neural, muller2006off}.
Given independence of the source of demonstration data, BC is not restricted to imitate human experts and can be applied with automatically generated trajectories \citep{2016arXiv160907910P}.

However, this application of naively trained supervised models \emph{in a non-iid scenario} comes with additional challenges as performed actions affect future input data. Small errors lead to the data distribution diverging from the training data, which commonly focuses on states along optimal trajectories, a phenomenon known as \emph{covariate shift}. Model performance degrades, potentially shifting the input data even further from our training data distribution, causing a \emph{compounding of errors} \citep{ross2010reduction}. To prevent this result, we need to learn how to recover from suboptimal states, which are usually not part of given expert demonstrations. \citep{bojarski2016end} addresses the problem in the context of lane keeping by synthetically generating off lane-centre states with additional cameras on the sides of the vehicle with corrected steering manoeuvres. One of the most common approaches is presented by DAgger \citep{ross2010reduction} and extensions, which collect additional expert supervision during application of the model - however this leads to increased efforts for providing supervision \citep{laskey2016shiv, kahn2017plato, laskey2017dart}. 
Finally, this type of end-to-end modelling is limited in terms of interpretability, and can rely on larger amounts of training data than modular or abstracted approaches \citep{2016arXiv160406915S}.

\textbf{Inverse Reinforcement Learning} (IRL) presents another popular approach to address the problem of \emph{covariate shift} - by blending supervised learning with reinforcement learning (RL) or planning to learn robust models. IRL aims to infer expert preferences by optimising a reward function that generates agent behaviour similar to the expert demonstrations - instead of an imitating policy. In the context of probabilistic models, it can be understood as optimising a model that maximises the probability of the expert's trajectories \citep{ziebart2008maximum}. 

While BC only learns accurate behaviour for expert-visited states, IRL extends to states visited by the RL or planning step and learns corrective behaviour when diverging from the original trajectories. Furthermore, recent work directly utilises human domain knowledge \citep{Wulfmeier2017IJRR} to define behaviour for states not sufficiently encountered by either. However, IRL in comparison to BC relies on an accurate systems model to simulate behaviour or the possibility to sample on the real system. If both are not possible we can turn, in the context of mobile robotics, to another approach based on supervised learning: training supervised segmentation models for traversable terrain \citep{2016arXiv161001238B, thrun2006probabilistic}.

Though the IRL problem is underconstrained, as many reward functions are able to describe the same optimal behaviour, various approaches have introduced simple assumptions to address the degeneracy and derive efficient, practical solutions \citep{ziebart2008maximum, ratliff2006maximum, choi2011map}. 
Based on these, impressive successes of IRL include learning artistic flying manoeuvres for an RC helicopter \citep{abbeel_heli} and predicting future motion for traffic participants such as cars and pedestrians \citep{kretzschmar2016socially, ziebart_planning-based_2009, NIPS2015_5882}. 

A major benefit of IRL-based methods lies in the integration into existing systems. By applying methods to learn cost (negated reward) functions for existing motion planning systems \citep{ratliff_learning_2009, Wulfmeier2017IJRR, shiarlis2017rapidly}, we can directly integrate learned models into deployable systems, which can straightforwardly be tested and benchmarked against existing, hand-crafted cost functions. 

\begin{figure}
    \centering
    \begin{subfigure}[t]{0.31\textwidth}
        \includegraphics[width=\textwidth, clip, trim = {0 1.5cm 0 1.5cm}]{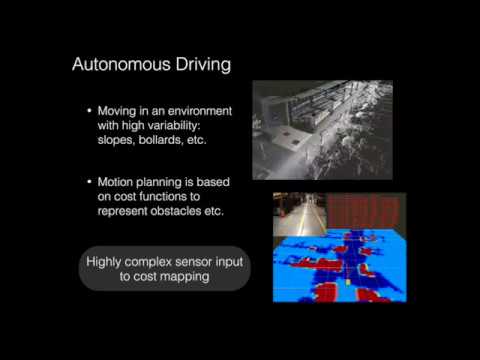}
        \caption{Cost-Function Learning via IRL \citep{Wulfmeier2017IJRR} (\href{https://www.youtube.com/watch?v=Sdfir_1T-UQ?rel=0}{link})}
    \end{subfigure} ~
    \begin{subfigure}[t]{0.31\textwidth}
        \includegraphics[width=\textwidth]{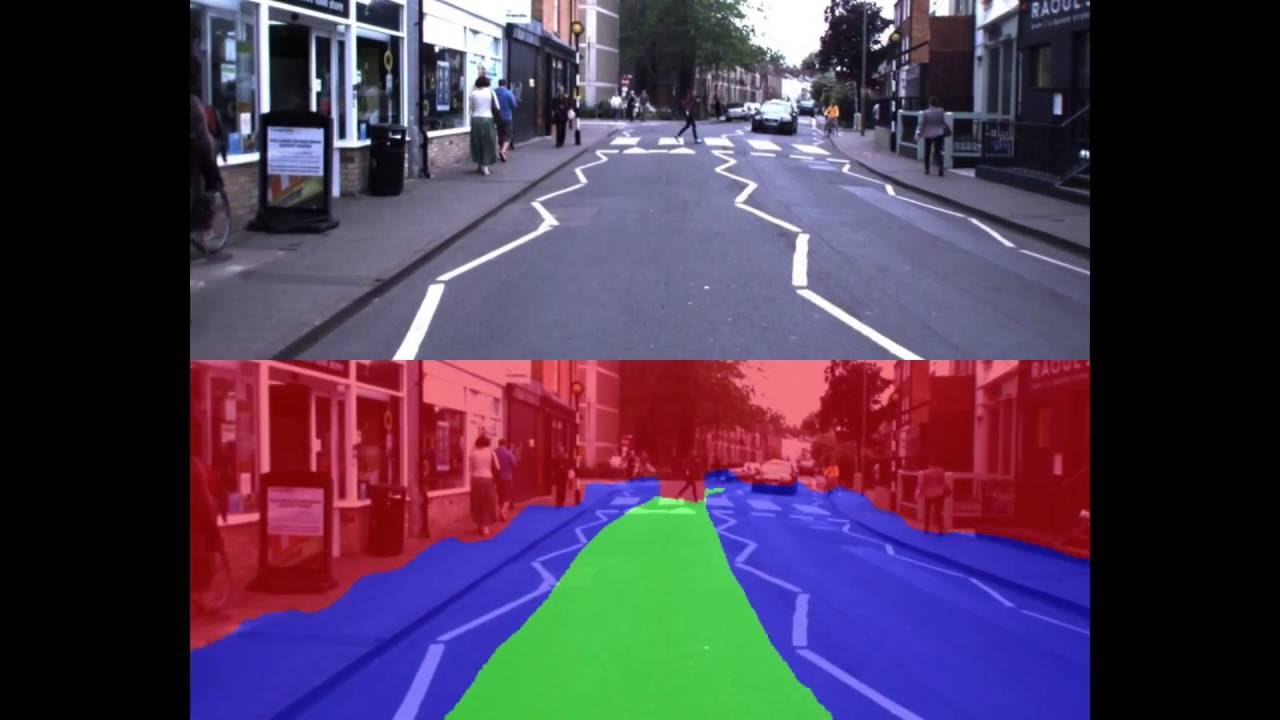}
        \caption{Terrain Classification \citep{2016arXiv161001238B} (\href{https://www.youtube.com/watch?v=rbZ8ck_1nZk?rel=0}{link})}
    \end{subfigure} ~
    \begin{subfigure}[t]{0.31\textwidth}
        \includegraphics[width=\textwidth, clip, trim = {0 1.5cm 0 1.5cm}]{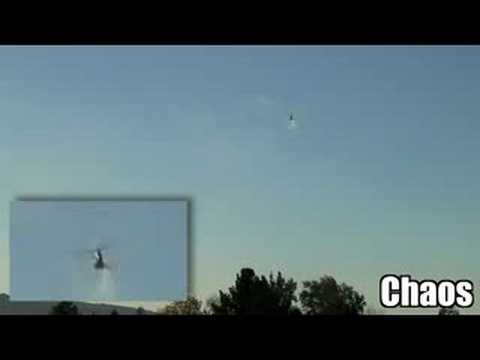}
        \caption{IRL for Flying \citep{abbeel_heli} (\href{https://www.youtube.com/watch?v=kN6ifrqwIMY?rel=0}{link})}
    \end{subfigure}
\end{figure}

When treating the robot control problem as part of a multi-agent scenario, we aim to optimise our actions not just for internal goals but as well for interaction and the internal goals of other agents. This approach gains relevance in cases when the other agents are represented by humans, as is the case in most robot applications. Research on \textbf{interpretability} of models has a long-standing history and recently gained increased attention based on the massive complexity and - more importantly - real world relevance of deep learning \citep{interpretability1, 2016arXiv160603490L, darpaexplainable}. The eminent aspect for control design is the interpretability of robot behaviour, urging us to \emph{act predictably} when directly interacting with humans \citep{humanlikedriving}. 
When planning motions in the direct vicinity of humans we benefit from providing non-verbal cues and human-like behaviour, enabling others to infer our driving style \citep{sadigh2016planning} and intentions \citep{PfeifferIROS2016, 2017arXiv170203465H} to ensure convenient and stressfree interaction. Broader surveys on legible behaviour for human robot interaction can be found in \citep{dragan2015legible, 2017arXiv170504226D}.

\subsection{Safety, Uncertainty and Introspection}\label{sec:safety}
The following sections present aspects which are disjoint of the modular pipeline structure addressed before and cover more general, cross-module concerns and potentials of learning from data in robotics.

Notably, relevant performance metrics in active, sequential decision making such as robotics and autonomous platforms differ from the metrics commonly  benchmarked for machine learning (such as classification accuracy, precision, recall, etc.). First, not all mistakes are equal: making a mistake confidently can be much more harmful than demonstrating \textbf{uncertainty} about the situation (e.g. the class of a pointcloud - pedestrian versus paper-bag). False negatives and false positives have massively different relevance for our modelling decisions. Second, we are able to act conservatively in the context of uncertainty and additionally probe the environment by collecting more data instead of forcing the model to make a confident prediction. Given the safety requirements for wide-scale application of autonomous vehicles \citep{waymosafety}, this approach is highly compelling. 

To determine the necessity of conservative behaviour, we depend on knowledge about the model's uncertainty for its predictions, which is commonly investigated as model \textbf{introspection}. \citep{grimmettuncertainty} investigates SVMs, GPs and a number of other popular classification models (pre deep learning) and empirically support the intuition that \emph{better introspection leads to improved decision making in the context of tasks such as autonomous driving or semantic map generation}. Furthermore, the authors indicate that commonly used detection metrics of precision and recall can be insufficient for describing model performance in safety-critical applications. 

Furthermore, commonly used \emph{pseudo probabilities} for introspecting model predictions, such as detection scores or softmax output, often do not suffice and supplemental uncertainty measures are required. 
\emph{Bayesian uncertainty} modelling for deep learning has found interest long before the current AI summer \citep{blundell2015weight, mackay1992practical, neal2012bayesian, yarinblog}, but recently received more spotlight thanks to the application of deep learning in safety-critical environments.
\citep{2017arXiv170304977K} investigates the use of \emph{aleatoric and epistemic} uncertainty metrics in deep learning, respectively for describing noise inherent in the observations and model uncertainty. Notably, while we cannot easily affect observation noise on the software side, we can reduce model uncertainty by collecting more data. Both metrics are suited for different purposes. While aleatoric uncertainty becomes highly relevant in large scale applications where epistemic uncertainty can be neglected, epistemic uncertainty can be employed to determine covariate shift between training and application data. In addition, this detection of \emph{novel} data can be addressed via generative models \citep{2017arXiv171108534W}.

\begin{figure}
    \centering
    \begin{subfigure}[t]{0.31\textwidth}
        \includegraphics[width=\textwidth]{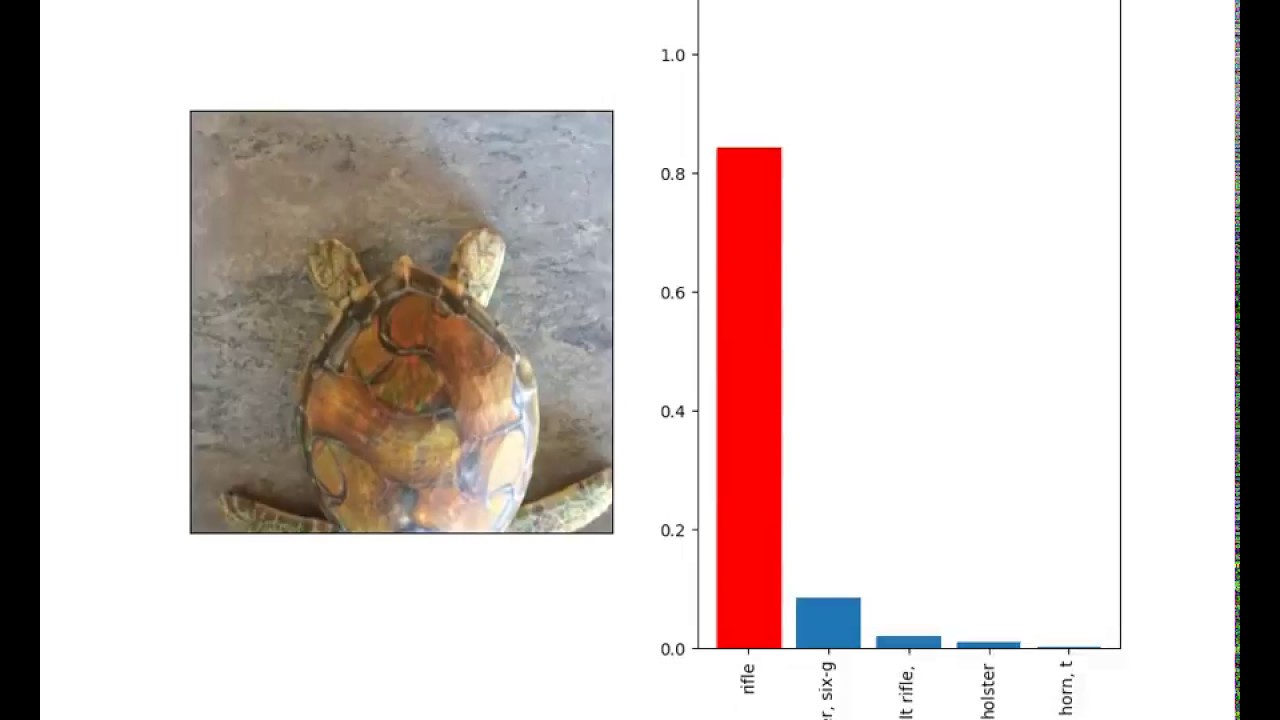}
        \caption{Synthesizing Robust Adversarial Examples \citep{2017arXiv170707397A} (\href{ https://www.youtube.com/watch?v=YXy6oX1iNoA?rel=0}{link})}
    \end{subfigure} ~
    \begin{subfigure}[t]{0.31\textwidth}
        \includegraphics[width=\textwidth]{Figures/ebn.jpg}
        \caption{Bayesian SegNet (precursor work to \citep{2017arXiv170304977K}) (\href{https://www.youtube.com/watch?v=Zu1BzW8fAjw?rel=0}{link})}
    \end{subfigure} ~
    \begin{subfigure}[t]{0.31\textwidth}
        \includegraphics[width=\textwidth]{Figures/ratslam.jpg}
        \caption{Microsoft CarSim \citep{carsim} (\href{https://www.youtube.com/watch?v=CauKo089zm0?rel=0}{link})}
    \end{subfigure}
\end{figure}

While the previously mentioned work aims at investigating and extending the capability of the model itself, a separate direction of research aims at \textbf{redundancy} and parallel streams of information to examine model predictions. 
Explicit introspection tools have been introduced for predicting the performance of perception modules \citep{gurau17}, the whole pipeline from perception to planning \citep{Daftry} and control \citep{2017arXiv171004459F}. 

Finally, \textbf{saliency detection} presents another essential approach to investigate model predictions, identifying input sections of high relevance. These visualisations are essentially obtained by determining the minimal input changes required to change model predictions \citep{2017arXiv170403296F, 2017arXiv170507857D, 2016arXiv161002391S, 2015arXiv151106581W}.

In order to accurately understand strengths and weaknesses of our system and improve on the latter, we depend on thorough testing and refinement of our systems. 
Training and testing systems in \textbf{simulation} enables us to repeat and vary edge cases. In essence, it enables us to generate multiple orders of magnitude more driving data \citep{waymosim, nytsim, ubersim} at different granularities. Various datasets and simulators are openly available for research \citep{dosovitskiy2017carla, carsim, richter2017playing, RosCVPR16, qiu2017unrealcv, 2016arXiv160506457G, 2016arXiv160801745S} with a more comprehensive list given in \citep{unrealcv}. While current simulators become increasingly accurate, the reality gap still persists and and emphasises the potential of related work on transfer learning and domain adaptation in Section \ref{sec:transfer}.

Just like every other software program, learning-based approaches have their \textbf{vulnerabilities} and can be fooled. New types of \emph{adversarial attacks} \citep{adversarial} - ways to mess with the input data to fool the model - as well as methods for defence have gained increased attention in the past years. Most impressively, recent work presents more general approaches and has shown adversarial examples with invariance with respect to 3D viewpoint \citep{2017arXiv170707397A} and attacked model \citep{moosavi2016universal}. Furthermore, \citep{2017arXiv170304730K} demonstrates the possibility of attacks on the training data set. Interesting work on defence against adversaries includes input transformations \citep{2017arXiv171100117G} and different encodings \citep{anonymous2018thermometer}.

\subsection{Knowledge Representation and Efficient Models}
\label{sec:knowledge}
Applications on mobile platforms and embedded systems have increased the demand for \textbf{computationally efficient} systems with reduced memory footprint aiming at on-chip rather than off-chip placement. The situation has lead to improvements of over an order of magnitude reduction in parameters, FLOPS, and the corresponding increase in possible frame-rate compared to previous state-of-the-art models with only limited reduction in accuracy \citep{2017arXiv171109224H, paszke2016enet, 2017arXiv170404861H}. 
Notable, the basic building blocks such as convolutions have been adapted via the introduction of asymmetric \citep{2015arXiv151200567S} and separable convolutions \citep{chollet2016xception} to increase parameter efficiency.

Work on model compression based on \emph{pruning, trained quantization and Huffman coding} was able to reduce the memory footprint of existing architectures \citep{2015arXiv151000149H, 2018arXiv180107365H}. Furthermore, instead of directly compressing the model, \emph{knowledge distillation} enables the training of smaller models via mimicking the predictions of large state-of-the-art architectures without their computational footprint \citep{bucilua2006model, hinton2015distilling}. The underlying intuition being that the logits extracted by the more powerful network include more information than the hard one-hot encodings, in particular about relations between classes. Recently, it has additionally been shown that distillation between networks of the same architecture can increase performance \citep{metalearn_furlanello}.

Finally, a role for learning can be found in \emph{reducing computation and time requirements} \citep{2016arXiv160907910P}. In this context, one perspective on AlphaGoZero \citep{silver2017mastering} covers the aspect of learning to imitate more expensive computations (here: MCTS), which enables the final, trained program to run faster and with lower computational requirements.
While the version of AlphaGo that bested Lee Sedol had an estimated \textbf{power consumption} of approximately 1 MW (50,000 times as much power as the amount of power required for a human brain), AlphaGoZero, which beat the previous version 100-0, uses an order of magnitude less compute.

\begin{figure}
    \centering
    \begin{subfigure}[t]{0.55\textwidth}
        \includegraphics[width=\textwidth]{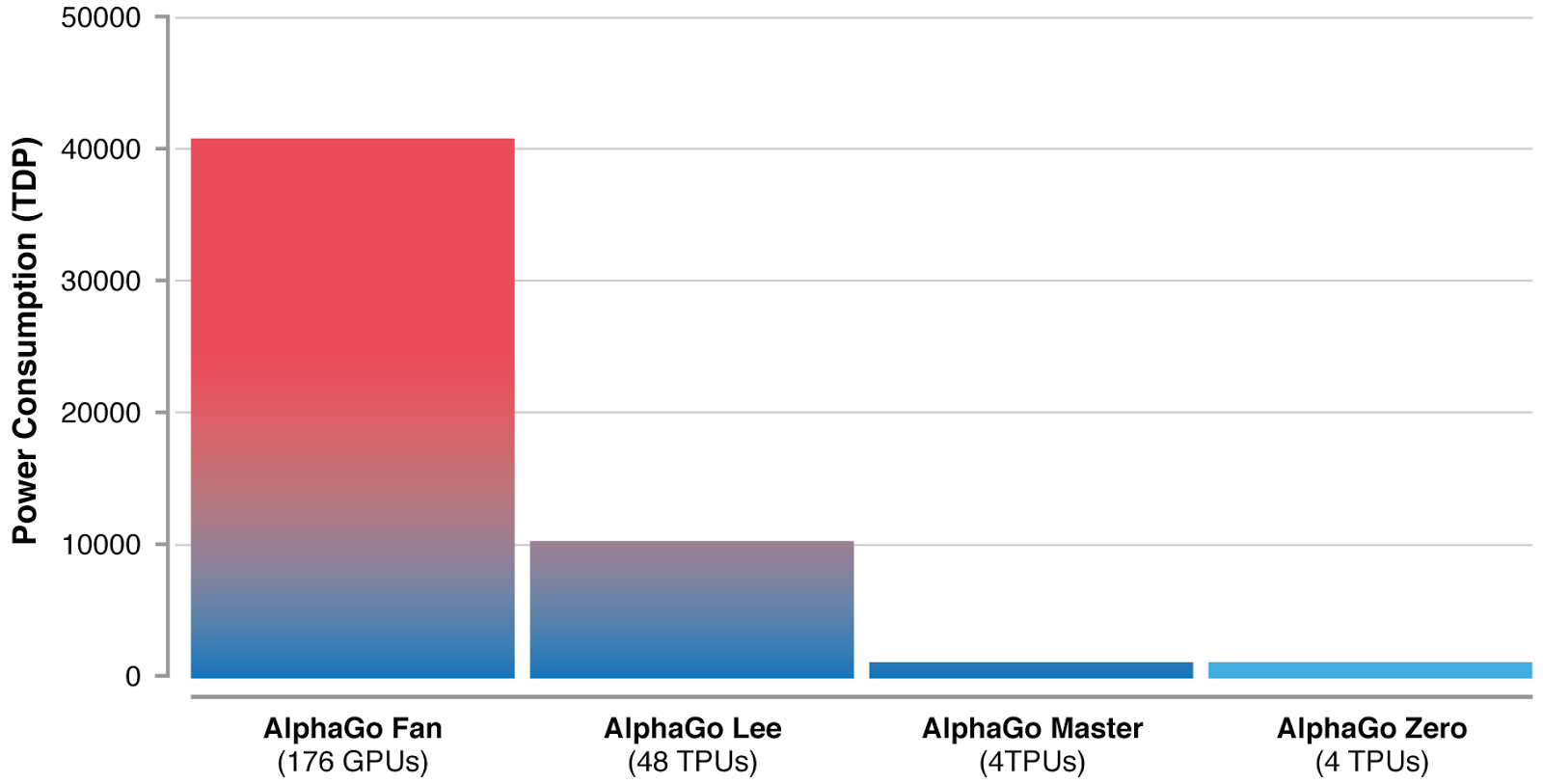}
        \caption{AlphaGo Power Consumption (source: businessinsider.com) (\href{http://uk.businessinsider.com/deepminds-alphago-ai-gets-alphago-zero-upgrade-2017-10}{link})}
    \end{subfigure} ~
    \begin{subfigure}[t]{0.40\textwidth}
        \includegraphics[width=\textwidth]{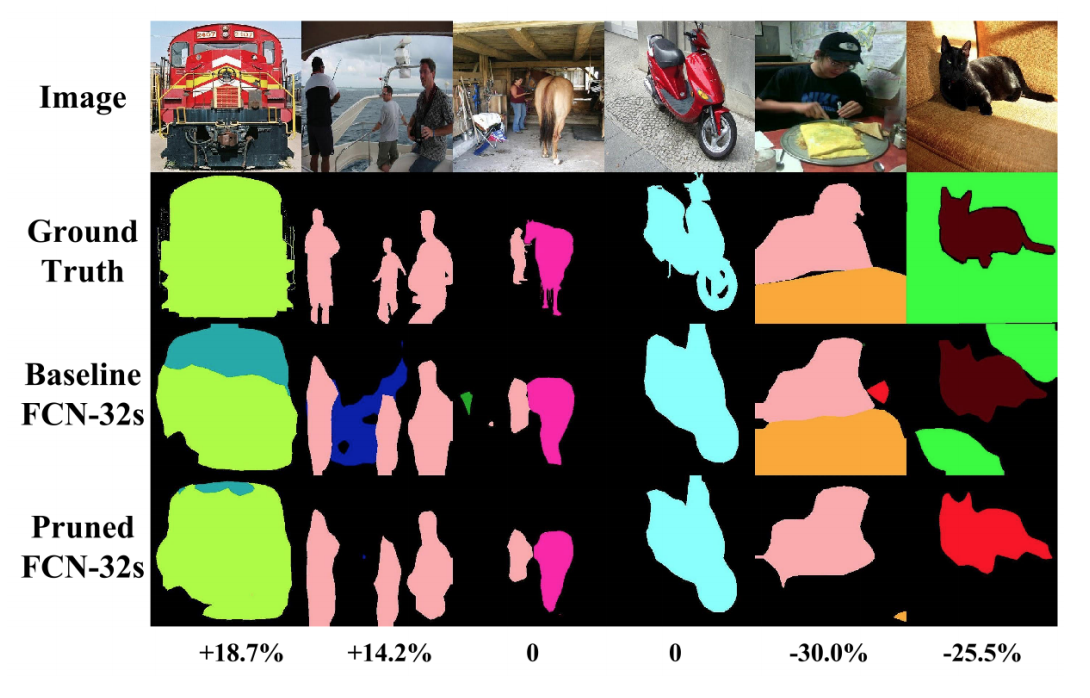}
        \caption{Learning to Prune in CNNs \citep{2018arXiv180107365H} (\href{https://www.youtube.com/watch?v=3yOZxmlBG3Y?rel=0}{link})}
    \end{subfigure}
\end{figure}

\subsection{Transfer, Multimodal, and Representation Learning}
\label{sec:transfer}
Robotics platforms perceive their environment through a multitude of different sensors. Learning can aid to combine and analyse this flood of information. 
Integrated training with \textbf{different sensing modalities} enables us to capture joint distributions, deploy with restricted access to our sensor setup \citep{ngiam2011multimodal, raomultimodal, valada2016deep} and increases robustness to sensor failure \citep{2017arXiv170510422L}.

The additional result of these high-throughput sensor setups is the generation of massive amounts of unsupervised data, exceeding our capability for manual dense annotation. 
Nonetheless, we can benefit from this overwhelming amount of data by splitting the problem in two: First, unsupervised learning of a representation from which we require less supervised data and second, provide supervision for determining the final mapping. By reducing the requirements for human annotation, \textbf{representation learning} (unsupervised learning) has significant potential, though has not yet found the same commercial success as supervised approaches. One of our challenges is that '\emph{we [often] don’t know what's a good representation}' \citep{bengio_interview}. Essentially, we're only able to \textbf{benchmark} in the context of surrogate tasks such as reconstruction accuracy or performance of subsequent classifier modules. 

Various approaches aim at finding relevant representations based on such proxy objectives; the aspects in common for most approaches is the prediction or verification of \textbf{structure - spatial and temporal}. 
Predicting spatial structure includes the relative location of image patches \citep{doersch2015unsupervised},  the order of shuffled image patches \citep{noroozi2016unsupervised}, image inpainting \citep{pathak2016context}, or employing various \emph{foundational [supervised] proxy 3D tasks} for learning a generic 3D representation \citep{zamir2016generic}.
The recent increase in compute capacity enables the extension of these ideas from the spatial to the temporal domain, thereby facilitating the use of temporal consistency and structure to learn representations from videos (e.g. \citep{unsupervisedviasequential}). Examples operate by verifying the temporal order of sequences \citep{2016arXiv160308561M}, predicting future frame representations \citep{vondrick2016anticipating}, or by predicting low-level motion-based clustering \citep{2016arXiv161206370P}.
Further detailed views on representation learning can be found in survey form \citep{2012arXiv12065538B}, in blog form on predictive versus representation learning \citep{grosserepresentationblog}, or in blog form on recent work and extensions to find training signals for RL \citep{unsupervisedblog}.

\begin{figure}
    \centering
    \begin{subfigure}[t]{0.31\textwidth}
        \includegraphics[width=\textwidth]{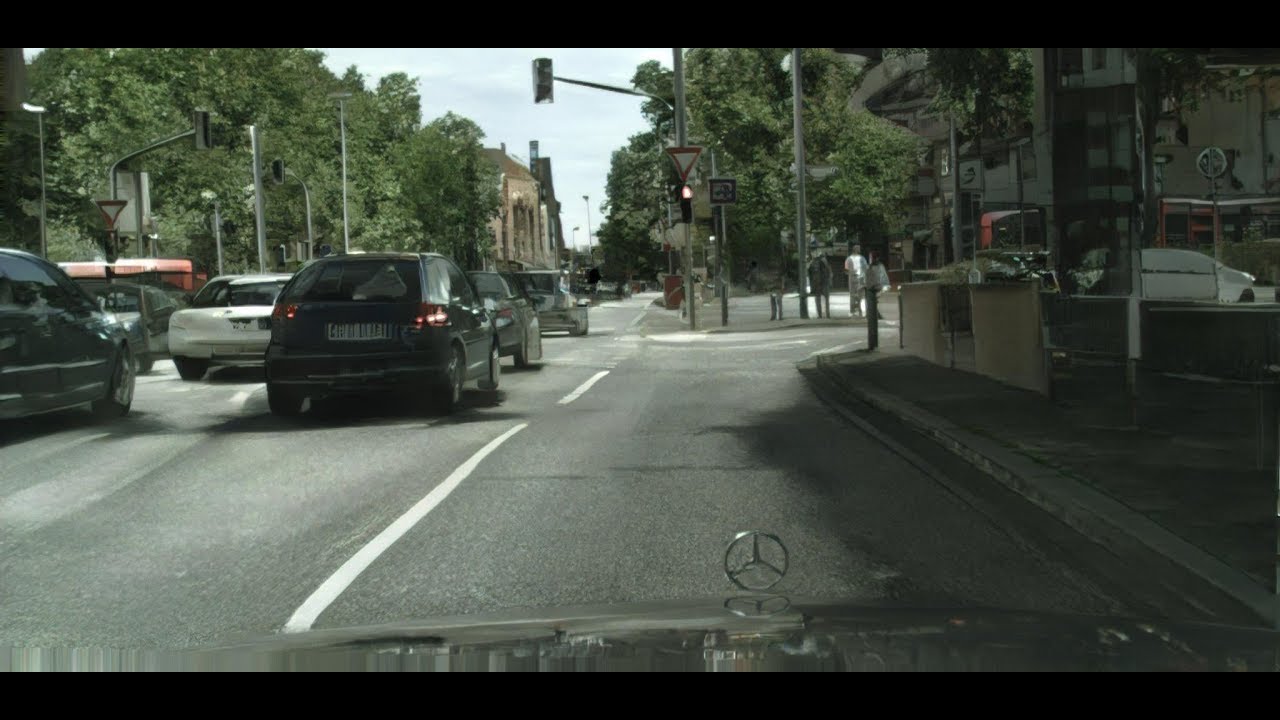}
        \caption{High-Resolution Image Synthesis and Semantic Manipulation with Conditional GANs \citep{2017arXiv171111585W} (\href{https://www.youtube.com/watch?v=3AIpPlzM_qs?rel=0}{link})}
    \end{subfigure} ~
    \begin{subfigure}[t]{0.31\textwidth}
        \includegraphics[width=\textwidth]{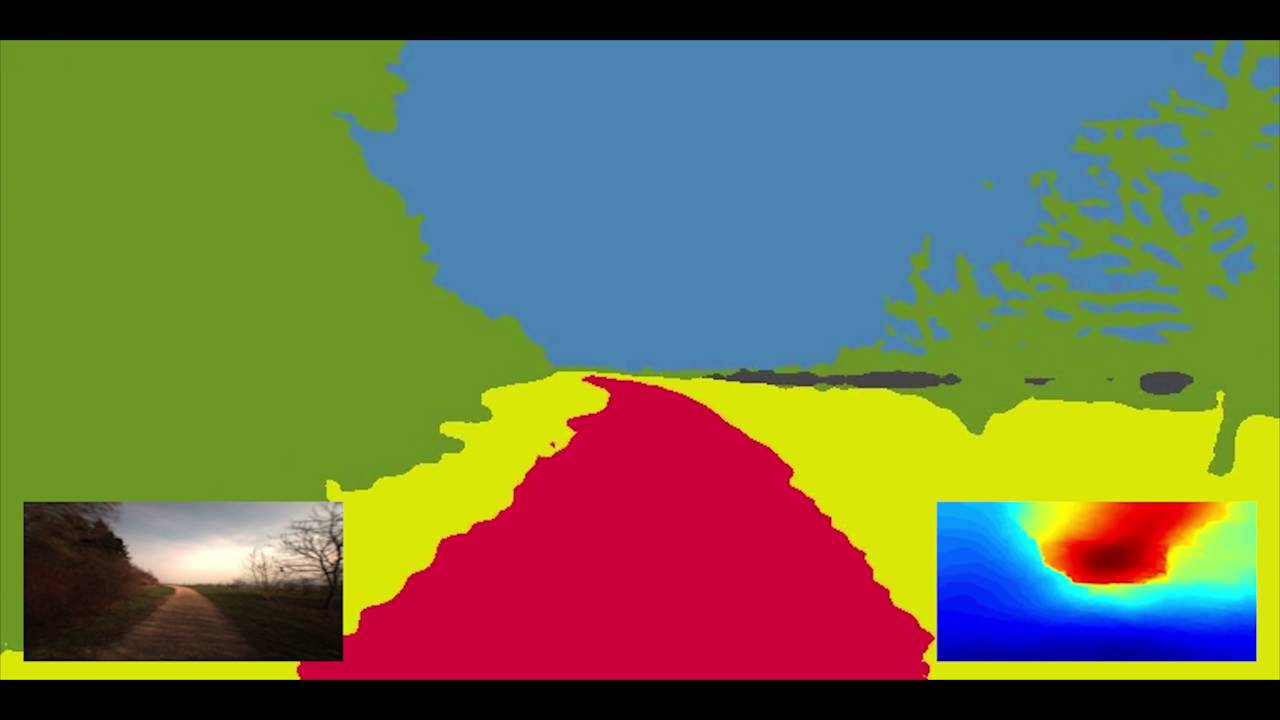}
        \caption{Deep Multispectral Semantic Scene Understanding \citep{valada2016deep} (\href{https://www.youtube.com/watch?v=GZvKAUAuKoA?rel=0}{link})}
    \end{subfigure} ~
    \begin{subfigure}[t]{0.31\textwidth}
        \includegraphics[width=\textwidth]{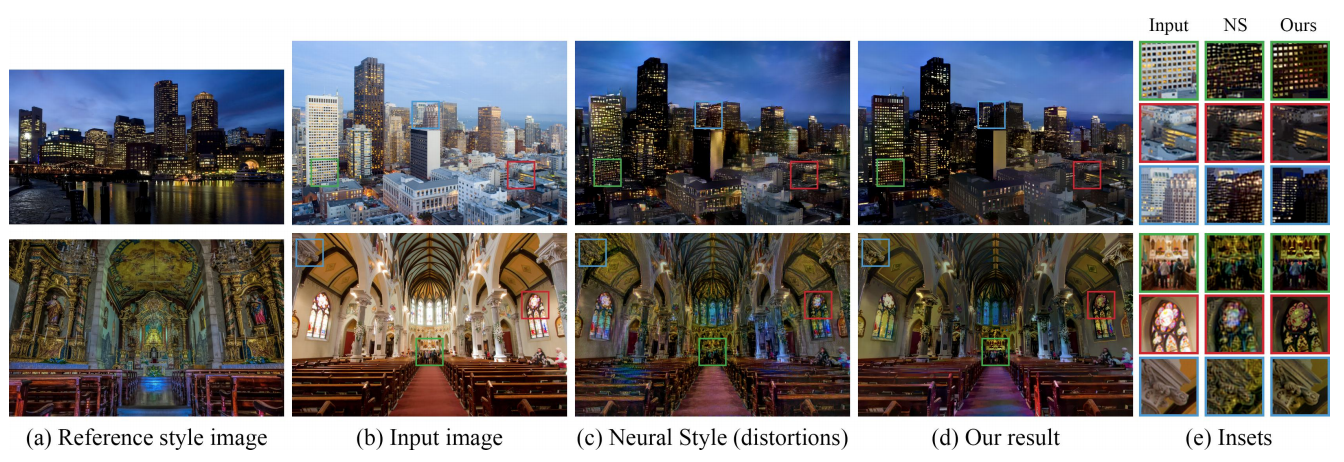}
        \caption{Deep Photo Style Transfer \citep{2017arXiv170307511L} (\href{https://www.youtube.com/watch?v=HTUxsrO-P_8?rel=0}{link})}
    \end{subfigure}
\end{figure}

One particularly relevant type of representation learning in this context is the field of unsupervised domain adaptation, which aims to induce \emph{domain invariant embeddings} such as to increase the performance for a task in domains without annotated data. The metric for evaluation is clearly defined in this case by the supervised task. 
Similar to various approaches emphasised in this review, the benefit of domain-invariant representations lies in \emph{addressing covariate shift}, the encounter of data outside our supervised training distribution during deployment of the model. Essentially, while acquiring widespread training and validation data is most beneficial \citep{mobileye}, the effort is often impractical based on expenses and the challenge of considering all potential conditions. 
Recent approaches aim at empowering domain adaptation methods via deep models \citep{Ganin2016, wulfmeier2017addressing, 2017arXiv170907857B, tzeng2015towards, csurka2017domain}; furthermore extending to adaptation in continually changing environments \citep{2017arXiv171207436W}.
Additionally to domain adaptation in the feature space of a model, transforming images between different domains has become a promising direction \citep{2017arXiv170907857B, 2017arXiv170310593Z, 2017arXiv171109020C, 2017arXiv171111585W, huang2017arbitrary}. Of course, fine tuning of pretrained models \citep{yosinski2014transferable, 2013arXiv13101531D} from different domains or different tasks remains nearly \citep{matl} always helpful if small amounts of supervised data are available for the application domain.

\section{Challenges and Potentials}
\label{sec:challenges}
After the review of current work on learning in different modules as well as at their intersections, this section concludes with promising directions and potentials as well as some challenges ahead.

\subsection{Bottom-Up}\label{sec:bottomup}
The principal strength of learning lies in applications in direct, reactive \textbf{perception} problems including classification, detection, segmentation, and related tasks. State-of-the-art models provide high accuracy with comparably little encoded structure (other than e.g. convolutions).

While there is no strong foundation for believing that \textbf{localisation} tasks are inherently much harder (or easier) to learn than e.g. object detection, geometric approaches simply provide a much stronger baseline in this field\citep{kitti, zhu2017image}.
Traditionally, CV addresses localisation very structured by utilising geometric knowledge about the mathematical rules underlying particular sub-problems (triangulation, homography estimation, etc). Instead of aiming to formalise rules for complete mappings from images to relative or absolute poses, odometry or SLAM systems extract and match features across sequences or between current perception and constructed map (while constructing said map), utilising prior knowledge about geometric properties, and improving pose graphs on the back-end, e.g. via filtering or bundle adjustment \citep{slam}. 
However, the first generation of learning-based methods addressed the problem with little to no structure, learning end-to-end mappings and based on pre-existing, annotated datasets (e.g. \citep{kendall2015posenet}). 
While given infinite data, covering the complete distribution of interest, and a flexible enough model, this approach can perform really well, the reality is often more constrained.

By combining both directions, the strength of geometric methods can have a substantial contribution towards more robust and reliable approaches. 
Various aspects of the task can be solved to perfect accuracy given geometric knowledge, while other aspects are affected by the real-world noise and can benefit via learning from data. 
Components that are likely to be focus on improvement via machine learning in future work include feature extraction e.g. for loop closure detection and re-localization or better point descriptors for sparse SLAM methods. Finally, deep learning can greatly improve the quality of map semantics - i.e. going beyond poses or pointclouds to a more complete understanding of the types and functionalities of objects or regions in the map. One particularly relevant direction lies in improved robustness (for example through better handling of dynamic objects and environmental changes \citep{2017arXiv171106623B}).  Learning will additionally be beneficial when the assumptions underlying traditional approaches are invalid or we require faster methods \citep{kendall2017geometric}.
Back at ICCV 2015, the question if deep learning would replace geometric CV for SLAM might have been received with significant scepticism \citep{slammatters} (and might still be), however most research has actually not tried to replace geometry but instead to enhance and augment, only learning parts of the system where cannot provide exact prior structure.

Commonly, the accuracy of perception and localisation systems can represent a \textbf{bottleneck} for overall performance and safety of a system as they provide the foundation for the rest of the pipeline, emphasising the relevance of even minute improvements. 
Part of the direct \emph{consumers} of their output are the \textbf{tracking and prediction} modules. 
Traditional tracking approaches (e.g. EKF, UKF) often provide reasonable solutions for the space of robotics. These methods are straightforward to implement and fully suffice as long as tracking itself does not become the bottleneck. 
Applications in more complex, cluttered environment, where data association becomes more challenging, represent a prime example of the benefits of learning for appearance-based tracking \citep{kristan2016visual, kristan2015visual, 2017arXiv170406036V, bertinetto2016fully}. Furthermore, densely populated scenarios often lead to more intricate interactions. Learned elaborate interactive motion models (in Section \ref{sec:tracking}) enable us to predict motion with higher accuracy in these environments and can be incorporated into existing trackers. 

Traditionally, \textbf{motion planning}, at least for deployed platforms, has been one of the fields more resistant to learning-based approaches (in particular in safety-critical applications). Planning approaches represent structured procedures for reasoning \citep{lavalle1998rapidly, pivtoraiko2005efficient}, utilising knowledge about e.g. kinematic and dynamic constraints, as well as geometric extension of platform and obstacles. 
As with localisation, parts of the planning computation are accurately modelled via known structure and equations (e.g. collision checking). 
In this context, learning focuses on more intuitive aspects, e.g. in improving prediction in interactive scenarios - such as highway lane merging - which requires to predict the reaction of other cars to potential actions. In essence, the more interactive and intuitive parts of driving, which are less governed by strict, easy-to-define rules, present opportunities for learning from data, where these forms of \emph{common sense} and \emph{intuition} are too complex for manual rules \citep{davis2015commonsense}.
Recent work on imitation learning for driving for example outsources high-level planning and takes additional commands as input \citep{2017arXiv171002410C} to focus learning on what is more straightforward to learn. This approach does not learn to plan but essentially a reactive controller (based on raw images) and presents another example for merging learning with existing systems. 

Route planning, as topological, high-level planning process, is commonly addressed via graph search (A* and related approaches). While there has been research on learning these kinds of programs as end-to-end approach in limited scenarios \citep{diffneuralcomputer, 2015arXiv151104834N}, given current applications, the existing algorithms do not represent a bottleneck. However, it can be expected that the costs associated with edges and nodes for the route graph are well-suited for estimation from data.  

Focusing on the incorporation of learning for planning and control into existing modular software pipelines, another principal application lies in the characterisation of traversability and obstacles as well as the prediction of the reactions of other traffic participants. Hand-crafted cost functions for different kinds of terrains are commonly designed to help bridge the gap between perception and action and reduce the complexity of our environment representation to focus on the aspects we care about. Learning cost functions for \emph{driving like human experts} (as e.g. in Section \ref{sec:planning}) is addressed via a sub-field of learning from demonstration \citep{shiarlis2017rapidly, Wulfmeier2017IJRR}. Furthermore, similar techniques can be applied to learn to predict reactive behaviours - interweaving planning and prediction models for dynamic environments \citep{PfeifferIROS2016}. Finally, in addition to manually defined cost functions, planning and reasoning systems commonly include many other heuristics, parameters determined during deployment to work in tested scenarios. Here, learning can play a major role in determining general rules on how to turn these knobs to adapt to new environments or different user preferences.

\begin{figure}
    \centering
    \begin{subfigure}[t]{0.45\textwidth}
        \includegraphics[width=\textwidth]{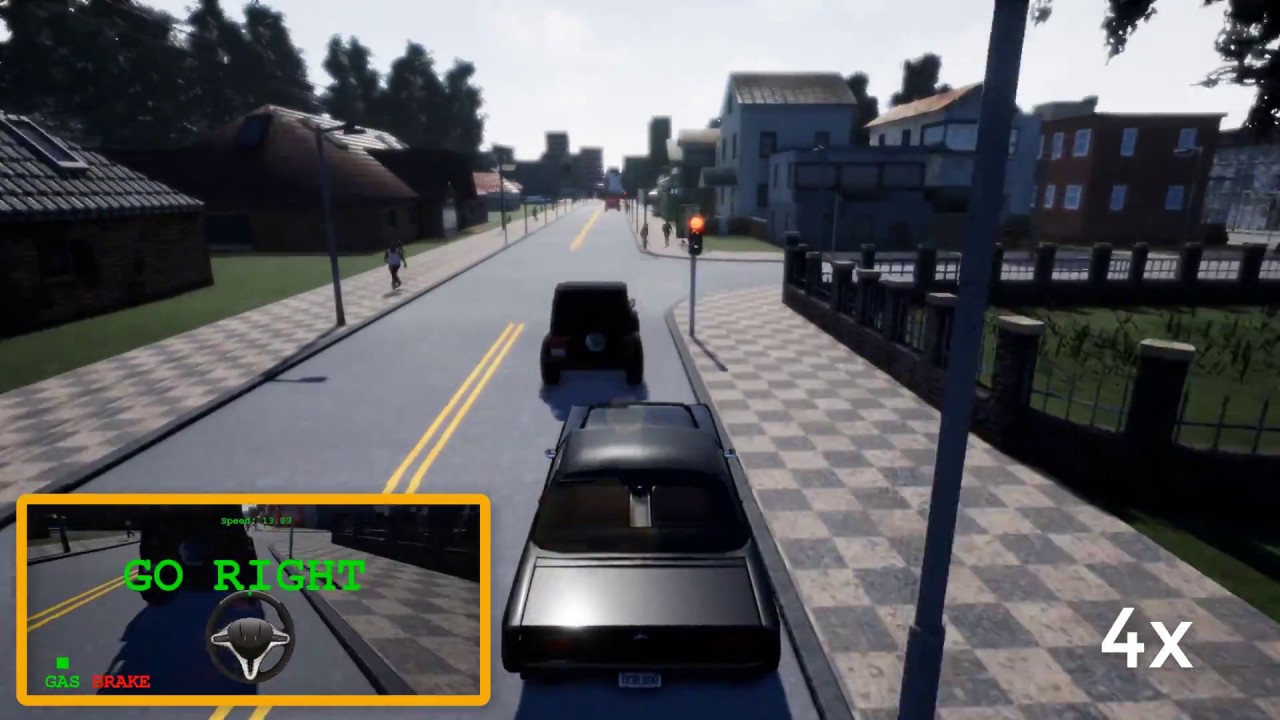}
        \caption{Conditional Imitation Learning \citep{2017arXiv171002410C} (\href{https://www.youtube.com/watch?v=cFtnflNe5fM?rel=0}{link})}
    \end{subfigure} ~
    \begin{subfigure}[t]{0.45\textwidth}
        \includegraphics[width=\textwidth]{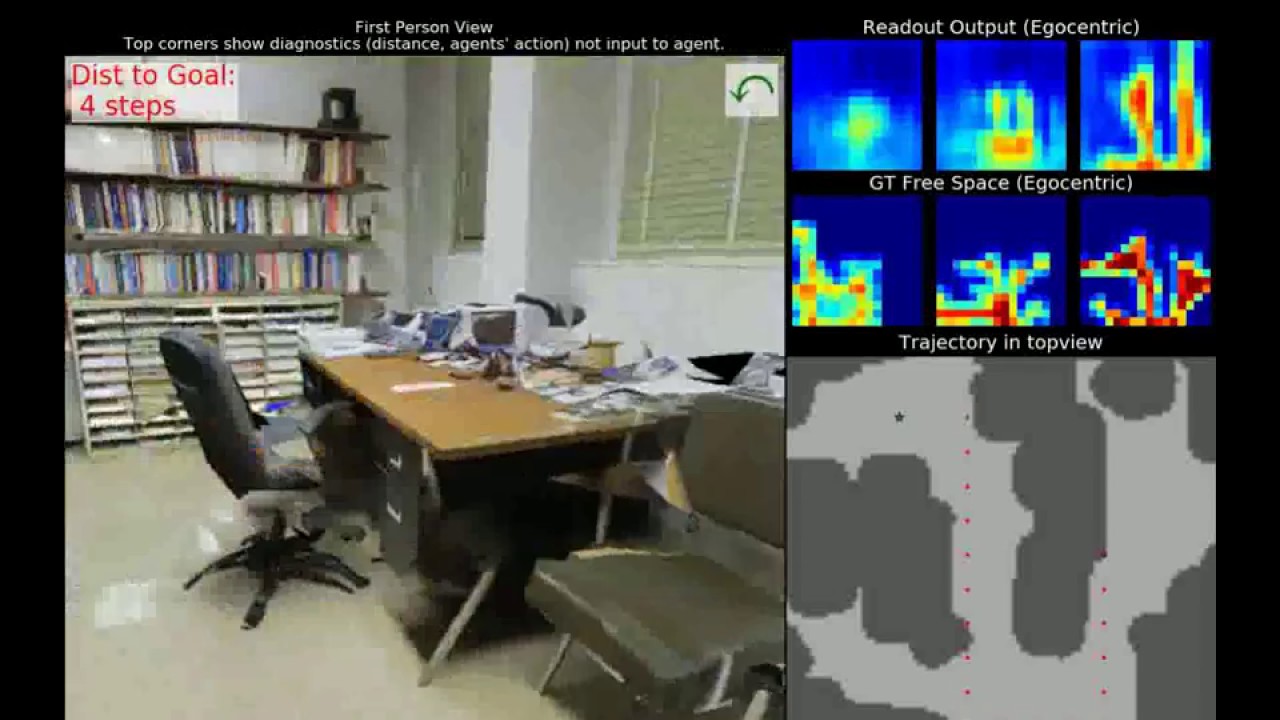}
        \caption{Cognitive Mapping and Planning \citep{2017arXiv170203920G} (\href{https://www.youtube.com/watch?v=BNmz3xBtcJ8?rel=0}{link})}
    \end{subfigure}
\end{figure}

In the context of safety-critical applications, learning is suitable for the generation of parallel systems for \textbf{redundancy} and additional checks, modules that replicate functionality and enable the second-guessing of decisions. Optimally, the application of multiple redundant modules is not restricted to verification but can culminate in a framework for \textbf{learning from disagreement}, where disagreements between modules are not only detected but feed back into the optimisation of the overall system, such as through adaptation of a module's uncertainty for future predictions.

A recent example for this type of framework is given by Pei et al \citep{pei2017deepxplore}. The authors devise an approach for sets of networks, retraining the modules that disagree with majority decisions. 
The underlying assumption, that the majority will always be right, is critical for success of the approach. Their training process aims at balancing two objectives: maximise the number of active neurons and trigger as many conflicts between the modules as possible. This objective is interestingly similar to basic ideas from software testing aiming to maximise code coverage. Variations of the idea aiming at adapting uncertainties and taking module uncertainty into account for a weighted majority represent promising further directions. 
Furthermore, it will be generally beneficial to address the transfer of various other concepts and paradigms which have been demonstrated successful and indispensable for software engineering.

\subsection{Top-Down}
\label{sec:topdown}
Machine learning (in particular deep learning) has the capability to extract rules from massive amounts of data and the benefit of high flexibility: merging of arbitrary objectives, Lego-like capabilities for the reuse and combination of models.
On the other hand, we have accurate mathematical formulations about the underlying math and programs to solve specific sub-problems e.g. for localisation and planning. 
Modern deep learning improves the ease for the integration of fixed and learned modules, enabling us to be standing on the shoulders of giants from both fields and build on known solutions. 

\pagebreak

While potentially trained as independent modules, the overall trajectory goes towards combinations which can be optimised as complete system via deterministic gradients as well as - if required - various stochastic gradient estimators (REINFORCE (or likelihood-ratio) trick \citep{williams1992simple}, evolution strategies \citep{back1996evolutionary, salimans2017evolution}, continuous relaxations such as Gumbel-Softmax or Concrete distribution and extensions \citep{maddison2016concrete, jang2016categorical, tucker2017rebar} ).

Combinations of both approaches can provide significant advantages via \textbf{redundant} systems and often \textbf{complementary} properties. 
In essence, we aim to take \emph{the best of both worlds} when merging two systems; similarly to how automation via ML aims to adapt and enhance job responsibilities \citep{humanplusai} by addressing tasks complementary to human strengths.
Ongoing directions include \textbf{optimising input data or correcting the output} of traditional programs. Examples for input improvement include learning image enhancement networks for traditional visual odometry methods \citep{2017arXiv170701274G}; output refinement includes pose correction updates for visual localisation \citep{2017arXiv170903128P} and refining dense reconstructions \citep{2018arXiv180109128T} as well as hand-crafted cost maps for motion planning \citep{Wulfmeier2017IJRR}. 

Similarly, pure learning-based approaches benefit from \textbf{incorporating prior knowledge} about the underlying structure: implicit and explicit translation invariance  \citep{lecun1989generalization, witten2016data}, \emph{objectness} \citep{2016arXiv160602378B}, temporal structure \citep{hochreiter1997long}, planning procedures such as value iteration \citep{2016arXiv160202867T}, further geometric properties \citep{2016arXiv160707405H, jaderberg2015spatial}, structure that encourages SLAM-like computations \citep{2017arXiv170609520Z} and access to SLAM - location and map - information for reinforcement learning \citep{2016arXiv161200380B}. 
Notably, the incorporation of structure can, under specific circumstances, even help when the incorporated models are inaccurate \citep{2017arXiv170706203W}.

The combination of prior geometric knowledge and the flexibility of learning enables the reformulation of geometric properties to create \textbf{self-supervised objectives}. 
Examples are given by \citep{2017arXiv170407804V, se3posenets} which utilise predictions for depth, segmentation masks and poses to differentiably warp frames in time to match image sections. In addition to training with externally supervised labels these approaches enable self-supervised learning e.g. via reprojection photometric errors. 
Lastly, the use of data augmentation represents the application of prior knowledge, structuring our wanted invariances via randomising the relevant aspects in our training data. 
A related survey on limits and potentials of deep learning in robotics can be found in \citep{2018arXiv180406557S}.

When moving from manually defining features and computations to designing the most efficient structure to learn into, one question arises naturally: why not \textbf{learn everything} (the architecture \citep{2018arXiv180201548R, 2017arXiv170805344B}, the optimiser \citep{2016arXiv161105763W, 2016arXiv160604474A}, or complete programs \citep{2018arXiv180202353K}). However, required investments in data hygiene and annotation for many applications with potential for real world impact often render it more efficient, in terms of human effort, to port our prior knowledge into algorithmic structure. 
Leslie Kaelbling formulated this well during a panel session at CoRL2017: 'What structure can we build in that does not obstruct learning?'. The point being twofold, with respect to model and the optimisation procedure. 
If structure is a \emph{necessary good} or \emph{necessary evil} might be up to discussion \citep{lecun_manning, innatemachinery, Georgeeaag2612, 2017arXiv171009829S}, but for now, practically, it is \textbf{necessary} as well as are the advantages of learning.

Building autonomous platforms, like addressing any other sufficiently complex and versatile software problem, results in a significant effort for \textbf{systems engineering and iterative testing and refinement}. The relative emphasis on learning or traditional programming blocks narrows down to the required effort and efficiency when creating reliable, safe and generalisable systems with either approach as well as the potential benefits of combination. 

\begin{figure}
    \centering
    \begin{subfigure}[t]{0.31\textwidth}
        \includegraphics[width=\textwidth]{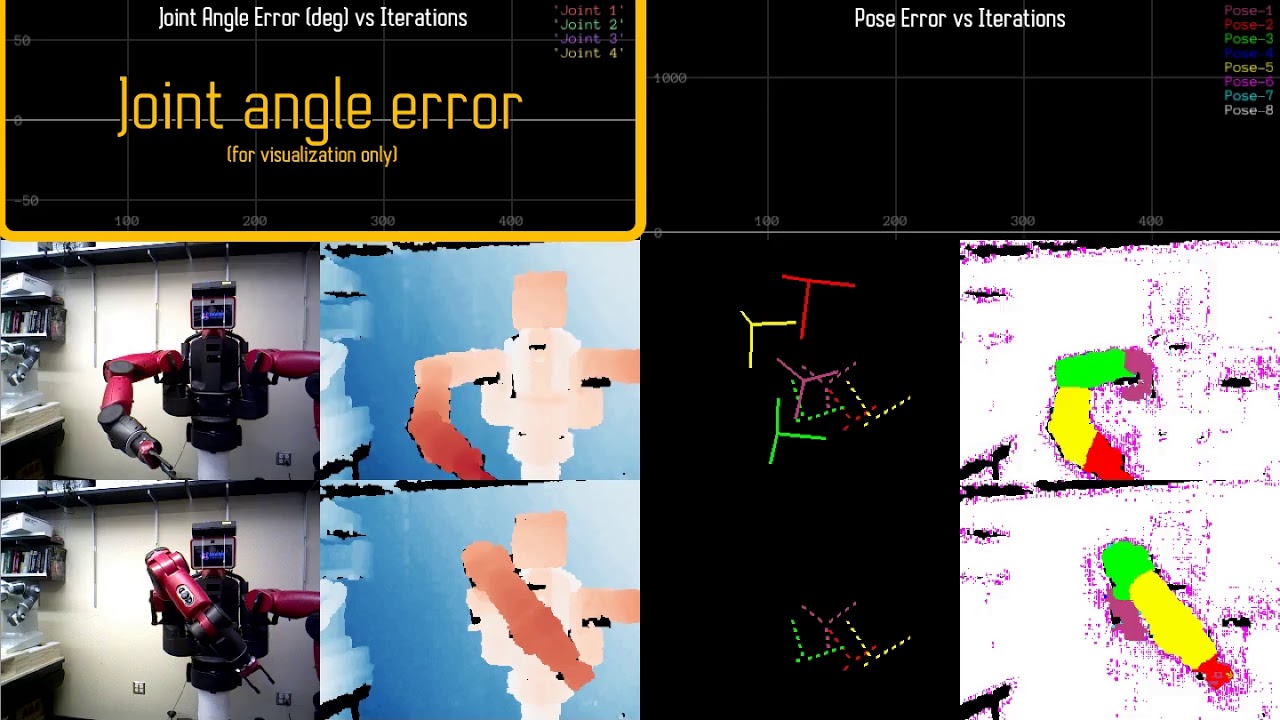}
        \caption{ SE3-Pose-Nets: Structured Deep Dynamics Models for Visuomotor Planning \citep{se3posenets} (\href{https://www.youtube.com/watch?v=MtMbv-LAnOM?rel=0}{link})}
    \end{subfigure} ~
    \begin{subfigure}[t]{0.31\textwidth}
        \includegraphics[width=\textwidth]{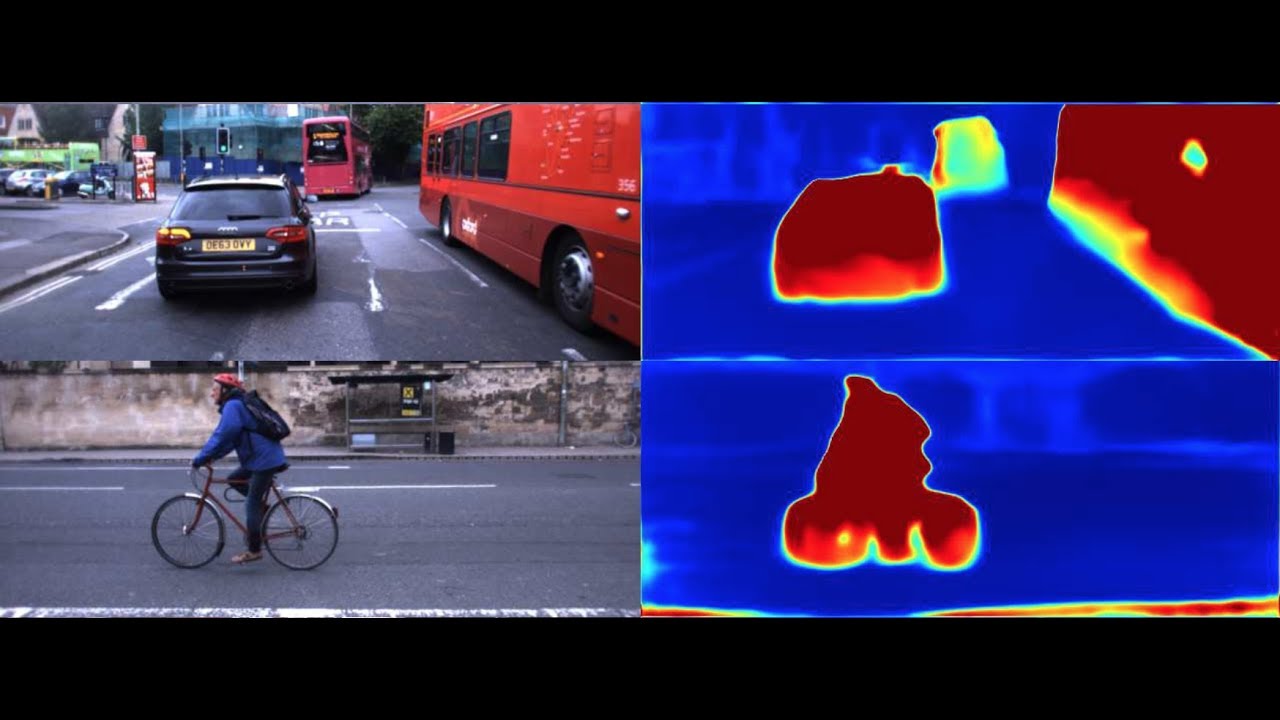}
        \caption{Driven to Distraction - deep learning for filtering dynamic objects \citep{2017arXiv171106623B} (\href{https://www.youtube.com/watch?v=ebIrBn_nc-k?rel=0}{link})}
    \end{subfigure} ~
    \begin{subfigure}[t]{0.31\textwidth}
        \includegraphics[width=\textwidth]{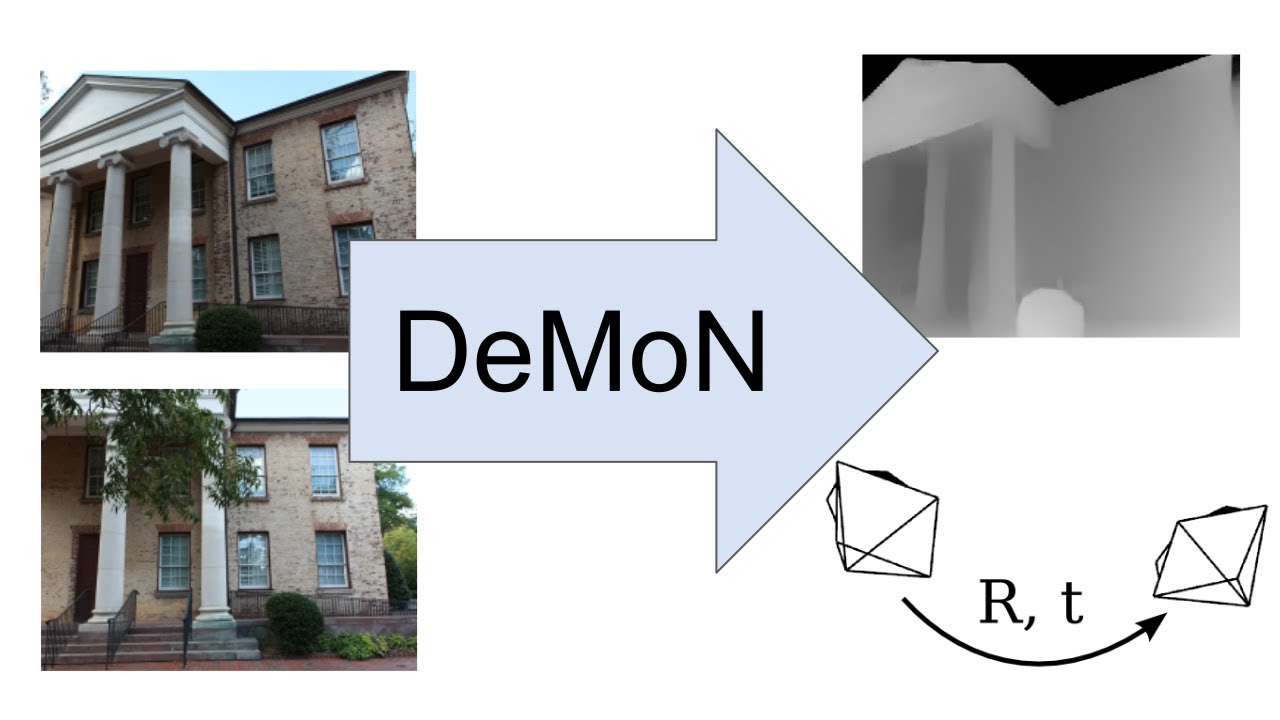}
        \caption{Depth and Motion Network for Learning Monocular Stereo \citep{2016arXiv161202401U} (\href{https://www.youtube.com/watch?v=Rat9-nyVd2s?rel=0}{link})}
    \end{subfigure}
\end{figure}

\pagebreak

\section*{Acknowledgements}

Over the last few years, discussions and collaborations directly and indirectly helped in shaping this article; with many colleagues and friends at the Oxford Robotics Institute, DeepMind and at BAIR. 
With respect to this review, I'm particularly thankful to Alex Bewley and Ankur Handa for thorough reading and feedback on earlier drafts.

I had the privilege of participating in a small percentage of the referenced work, none of which could have been done without an amazing group of collaborators including Ingmar Posner, Dushyant Rao, Alex Bewley, Dominic Zeng Wang, Peter Ondruska, Dave Held, Carlos Florensa, Pieter Abbeel, Kyriacos Shiarli, Sasha Salter and Shimon Whiteson.

Even though not involved in chats about this review or previous publications, many others have been involved indirectly in shaping ideas: Julie Dequaire, Corina Gurau, Jeff Hawke, Martin Engelcke, Adam Kosiorek, Fabian Fuchs, Oliver Groth, Abishek Gupta, Martin Riedmiller, Raia Hadsell, Jonas Buchli, Larry Zitnick, Anca Dragan, Nathan Benaich, Shubho Sengupta and many others. However, it is impossible to provide a complete list here.

\catcode`\_=12
\bibliographystyle{abbrv}
\bibliography{references}

\end{document}